\documentclass[10pt,twocolumn,letterpaper]{article}

\usepackage{cvpr}
\usepackage{times}
\usepackage{epsfig}
\usepackage{graphicx}
\usepackage{amsmath}
\usepackage{amssymb}
\usepackage{graphicx}
\usepackage{chngpage}
\usepackage{xcolor}
\usepackage{subfig}
\usepackage{booktabs}       
\usepackage{dblfnote}
\usepackage{dblfloatfix}
\usepackage{tablefootnote}
\usepackage{tabularx}
\usepackage{multicol}

\newcommand{\figref}[1]{Fig. \ref{#1}}
\newcommand{\tabref}[1]{Table \ref{#1}}
\newcommand{\secref}[1]{Sec. \ref{#1}}
\newcommand{\eqnref}[1]{Eqn. \ref{#1}}

\usepackage[pagebackref=true,breaklinks=true,letterpaper=true,colorlinks,bookmarks=false]{hyperref}

\cvprfinalcopy 

\def\cvprPaperID{3091} 

\ifcvprfinal\fi
\begin{document}

\title{Bilateral Cyclic Constraint and Adaptive Regularization \\ for Unsupervised Monocular Depth Prediction}

\author{Alex Wong\\
University of California, Los Angeles\\
{\tt\small alexw@cs.ucla.edu}
\and
Byung-Woo Hong\\
Chung-Ang University, Korea\\
{\tt\small hong@cau.ac.kr}\\
\and
Stefano Soatto\\
University of California, Los Angeles\\
{\tt\small soatto@cs.ucla.edu}
}

\maketitle

\begin{abstract}
Supervised learning methods to infer (hypothesize) depth of a scene from a single image require costly per-pixel ground-truth. We follow a geometric approach that exploits abundant stereo imagery to learn a model to hypothesize scene structure without direct supervision. Although we train a network with stereo pairs, we only require a single image at test time to hypothesize disparity or depth. We propose a novel objective function that exploits the bilateral cyclic relationship between the left and right disparities and we introduce an adaptive regularization scheme that allows the network to handle both the co-visible and occluded regions in a stereo pair. This process ultimately produces a model to generate hypotheses for the 3-dimensional structure of the scene as viewed in a single image. When used to generate a single (most probable) estimate of depth, our method outperforms state-of-the-art unsupervised monocular depth prediction methods on the KITTI benchmarks. We show that our method generalizes well by applying our models trained on KITTI to the Make3d dataset. \\
{\color{blue} Code and pre-trained models available at: \\ https://github.com/alexklwong/adareg-monodispnet}
\end{abstract}
\vspace{-0.0em}
\section{Introduction}
\vspace{-0.4em}
\label{introduction}
Estimating the 3-dimensional geometry of a scene is a fundamental problem in machine perception with a wide range of applications, including autonomous driving \cite{janai2017computer}, robotics \cite{lenz2015deep,stoyanov2010real}, pose-estimation \cite{shotton2013real} and scene object composition \cite{karsch2014automatic}. It is well-known that 3-d scene geometry can be recovered from multiple images of a scene taken from different viewpoints, including stereo, under suitable conditions. Under no conditions, however, is a single image sufficient to recover 3-d scene structure, unless prior knowledge is available on the shape of objects populating the scene. Even in such cases, metric information is lost in the projection, so at best we can use a single image to generate hypotheses, as opposed to  estimates, of scene geometry. 

Recent works \cite{chen2016single, eigen2014depth, laina2016deeper, liu2015deep, liu2016learning, xu2017multi, xu2018structured} sought to exploit such strong scene priors by using pixel-level depth annotation captured with a range sensor (e.g. depth camera, lidar) to regress depth from the RGB image. Cognizant of the intrinsic limitations of this endeavor, we exploit stereo imagery to train a network without ground-truth supervision for generating depth hypotheses, to be used as a reference for 3-d reconstruction. We evaluate our method against ground-truth depths via two benchmarks from the KITTI dataset \cite{geiger2012we} and show that it generalizes well by applying models trained on KITTI to Make3d \cite{saxena2009make3d}.

Rather than attempting to learn a prior by associating the raw-pixel values with depth, we recast depth estimation as an image reconstruction problem \cite{garg2016unsupervised, godard2017unsupervised} and exploit the epipolar geometry between images in a rectified stereo pair to train a deep fully convolutional network. Our network learns to predict the dense pixel correspondences (disparity field) between the stereo pair, despite only having seen one of them. Hence, our network implicitly learns the relative pose of the cameras used in training and hallucinates the existence of a second image taken from the same relative pose when given a single image during testing. From the disparity predictions, we can synthesize depth using the known focal length and baseline of the cameras used in training.     

While \cite{garg2016unsupervised, godard2017unsupervised, xie2016deep3d} follow a similar training scheme, \cite{xie2016deep3d} does not scale to high resolution, and \cite{garg2016unsupervised} uses a non-differentiable objectives. \cite{godard2017unsupervised} proposed using two uni-directional edge-aware disparity gradients and left-right disparity consistency as regularizers. However, edge-awareness should inform bidirectionally and left-right consistency suffers from occlusions and dis-occlusions. Moreover, regularity should not only be data-driven, but also model-driven. 

\noindent {\bf Our contributions} are three-fold: (i) A model-driven adaptive weighting scheme that is both space- and training-time varying and can be applied generically to regularizers. (ii) A bilateral consistency constraint that enforces the cyclic application of left and right disparity to be the identity. (iii) A two-branch decoder that specifically learns the features necessary to maximize data fidelity and utilizes such features to refine an initial prediction by enforcing regularity. We formulate our contributions as an objective function that, when realized even by a generic encoder-decoder, achieves state-of-the-art performance on two KITTI \cite{geiger2012we} benchmarks and exhibits generalizability to Make3d \cite{saxena2009make3d}.
\vspace{-0.0em}
\section{Related Works}
\vspace{-0.4em}
\label{related_works}
\noindent \textbf{Supervised Monocular Depth Estimation.} 
\cite{saxena2006learning} proposed a patch-based model that estimated a set of planes to explain each patch. The local estimates were then combined with Markov random fields (MRF) to estimate the global depth. Similarly,  \cite{hoiem2007recovering, karsch2012depth, konrad2013learning, saxena2009make3d} exploited local monocular features to make global predictions. However, local methods lack the global context needed to generate accurate depth estimates. \cite{liu2016learning} instead employed a convolutional neural network (CNN). \cite{ladicky2014pulling} further improved monocular methods by incorporating semantic cues into their model.

\cite{eigen2015predicting, eigen2014depth} introduced a two scale network to learn a representation from the images and regress depth directly. \cite{laina2016deeper} proposed a residual network with up-sampling modules to produce higher resolution depth maps. \cite{chen2016single} learned depth using crowd-sourced annotations of relative depths and \cite{fu2018deep} learned the ordinal relations of the scene using atrous spatial pyramid pooling. \cite{roy2016monocular} explored a patch-based approach using neural forests and \cite{kim2016unified, xu2017multi, xu2018structured} used conditional random fields (CRF) jointly with a CNN to estimate depth.  

\noindent \textbf{Unsupervised Monocular Depth Estimation.}
Recently, \cite{flynn2016deepstereo} introduced novel view synthesis by predicting pixel values based on interpolation from nearby images. \cite{xie2016deep3d} minimized an image reconstruction loss to hallucinate the existence of a right view of a stereo pair given the left by producing the distribution of disparities for each pixel.

\cite{garg2016unsupervised} trained a network for monocular depth prediction by reconstructing the right image of a stereo pair with the left and synthesizing disparity as an intermediate step. Yet, their image formation model is not fully differentiable, making their objective function difficult to optimize. Unsupervised methods \cite{godard2017unsupervised, patraucean2015spatio, zhou2016learning, zhou2016view} utilized a bilinear sampler modeled after the Spatial Transformer Network \cite{jaderberg2015spatial} to allow for a fully differentiable loss and end-to-end training of their respective networks. Specifically, \cite{godard2017unsupervised} used SSIM \cite{wang2004image} as a loss in addition to the image reconstruction loss. Also, \cite{godard2017unsupervised} predicted both left and right disparities and used them for regularization via a left-right consistency check along with an edge-aware smoothness term. \cite{aleotti2018generative} trains a Generative Adversarial Network (GAN) \cite{goodfellow2014generative} to constrain the output to reconstruct a realistic image to reduce the artifacts seen from stereo reconstruction.

Self-supervised methods \cite{mahjourian2018unsupervised, ummenhofer2017demon, zhou2017unsupervised, zou2018df} utilized a pose network to learn both ego-motion and depth from monocular videos sequences, while \cite{wang2018learning, yang2018deep} leveraged visual odometry from off-the-shelf methods \cite{engel2018direct, steinbrucker2011real} and \cite{fei2018geo} gravity as  supervisors. \cite{zhan2018unsupervised} followed both unsupervised and self-supervised paradigms by training multiple networks using stereo video streams and proposed a feature reconstruction loss. While additional supervision and data are used to improve predictions, \cite{godard2017unsupervised} still remains as the state-of-the-art in the unsupervised setting. Our method follows the unsupervised paradigm and we show that it not only outperforms \cite{godard2017unsupervised}, but also \cite{zhan2018unsupervised} who leveraged techniques from both unsupervised and self-supervised domains.

\noindent \textbf{Adaptive Regularization.} 
A number of computer vision problems can be formulated as energy minimization in a variational framework with a data fidelity term and a regularizer weighted by a fixed scalar. The solution found by the minimal energy involves a trade-off between data fidelity and regularization. Finding the optimal parameter for regularity is a long studied problem as \cite{galatsanos1992methods} explored methods to determine the regularization parameter in image de-noising, while \cite{nguyen2001efficient} used cross-validation as a selection criterion for the weight. Others \cite{godard2017unsupervised, wedel2009structure, werlberger2009anisotropic} used image gradients as cues for a data-driven weighting scheme. Recently, \cite{hong2017adaptiveinverse, hong2017adaptive} proposed that regularity should not only be data-driven, but also model driven. The amount of regularity imposed should adapt to the fitness of the model in relation to the data rather than being constant throughout the training process. 

We propose a novel objective function using bilateral cyclic consistency constraint along with a spatial and temporal varying regularization modulator. We show that despite using the fewer parameters than \cite{godard2017unsupervised}, we outperform \cite{godard2017unsupervised} and other unsupervised methods. We detail our loss function with adaptive regularization, in \secref{formulation}, present a two-branch decoder architecture in \secref{two_branch_decoder}, and specify hyper-parameters and data augmentation procedures used in \secref{implementation_details}. We evaluate our model on the KITTI 2015, KITTI Eigen Split, and Make3d benchmarks in \secref{experiments_results}. Lastly, we end with a discussion of our work in \secref{discussion}. 
\vspace{-0.0em}
\section{Method Formulation}
\vspace{-0.4em}
\label{formulation}

We learn a model to hypothesize or ``estimate'' the disparity field $d$ compatible with an image $I^0$ by exploiting the availability of stereo pairs $(I^0, I^1)$ during training. We then synthesize the depth $z=FB/d$ of the scene using the focal length $F$ and baseline $B$ during test time. Given $I^0$, we estimate a function $d \in \mathbb{R}_{+}$ that represents the disparity of $I^0$, which we formulate as a loss function $L$ (\eqnref{loss_function_eqn}), comprised of data terms and adaptive regularizers.

Our network, parameterized by $\omega$, takes a single image $I^0$ as input and estimates a function $d = f(I^0; \omega)$, where $d$ represents the disparity (which is monotonically related to inverse-depth) corresponding to $I^0$. We drive the training process with $I^1$, which is only used in the loss function, by a surrogate loss that minimizes the reprojection error of $I^0$ to $I^1$ and vice versa. 
We will refer to the disparity estimated by $L$ as $d^0$ and $d^1$ for $I^0$ and $I^1$, respectively. Interested readers may refer to Supplementary Materials (Supp. Mat.) for more details on our formulation. 
\begin{align}
\label{loss_function_eqn}
  	L = \underbrace{w_{ph}l_{ph}+w_{st}l_{st}}_{\text{data fidelity}}+\underbrace{w_{sm}l_{sm}+w_{bc}l_{bc}}_{\text{regularization}}
\end{align}
where each individual term $l$ will be described in the next sections and their weights $w$ in \secref{implementation_details}.

\vspace{-0.0em}
\subsection{Data Fidelity}
\vspace{-0.3em}
\label{data_unaries}
Our data fidelity terms seek to minimize the discrepancy between the observed stereo pair $(I^0, I^1)$ and their reconstructions $(\hat{I}^0, \hat{I}^1)$. We generate each $\hat{I}$ term by applying a 1-d horizontal disparity shift to $I$ at each position $x \in \Omega$: 
\begin{align}
\label{reprojection_eqn}
  	\hat{I}^0(x) = I^1(x-d^0(x)) \text{ \ and \ } \hat{I}^1(x) = I^0(x+d^1(x))
\end{align}
We do so by using a 1-d horizontal bilinear sampler modeled after the image sampler from the Spatial Transformer Network \cite{jaderberg2015spatial} -- instead of applying an affine transformation to activations, we warp an image to the domain of its stereo-counterpart using disparities. Our sampler is locally fully differentiable and each output pixel is the weighted sum of two (left and right) pixels.
We propose to minimize the reprojection residuals as a two-part loss, which measures the standard color constancy (photometric) and the difference in illumination, contrast and image quality (structural).

\noindent \textbf{Photometric loss.} We model the image formation process via a photometric loss $l_{ph}$, which measures the $L_1$ penalty of the reprojection residual for each $I$ and $\hat{I}$ on each channel at every position $x \in \Omega$ for ${s \in S \doteq \{0, 1\}}$ denoting the left and right images: 
\begin{equation}
\begin{aligned}  
\label{photometric_loss_eqn}
  	l_{ph} = \sum_{s \in S} \sum_{x \in \Omega} | I^s(x)-\hat{I}^s(x) |
\end{aligned}
\end{equation}

\noindent \textbf{Structural loss.} In order to make inference invariant to local illumination changes, we use a perceptual metric (SSIM) that discounts such variability. We apply SSIM ($\phi$) to image patches of size $3\times3$ at corresponding $x$ in $I$ and $\hat{I}$. Since two similar images give a SSIM score close to 1, we subtract 1 by the score to represent a distance:
\begin{align}  
\label{structural_loss_eqn}
  	l_{st} = \sum_{s \in S} \sum_{x \in \Omega} 1-(\phi(I^s(x), \hat{I}^s(x))
\end{align}

\begin{figure*}
    \centering
    \includegraphics[width=0.99\textwidth]
        {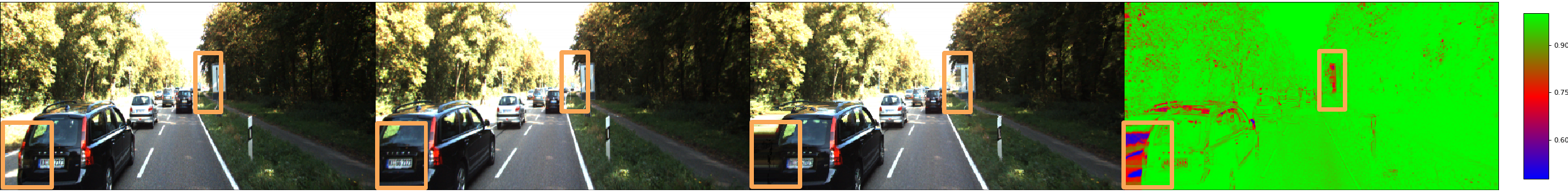}
    \vspace{-0.5em}
    \caption{Left to right: left image, right image, left reconstruction, adaptive weights. The adaptive weights reduce regularization at regions of high residual; hence, they discount dis-occlusions and occlusions as in the highlighted regions.}
    \vspace{-1.5em}
    \label{fig:adaptive_weights}
\end{figure*}

\vspace{-0.0em}
\subsection{Residual-Based Adaptive Weighting Scheme}
\vspace{-0.4em}
\label{residual_weights}
A point estimate $d$ can be obtained by maximizing the Bayesian criterion with a data fidelity term (energy) $\mathcal{D}(d)$ and a Bayesian or Tikhonov regularizer $\mathcal{R}(d)$ in the form:
\vspace{-0.4em}
\begin{align}  
\label{energy_function_eqn}
  	\mathcal{D}(d)+\alpha \mathcal{R}(d)
\end{align}
where the weight $\alpha$ is a pre-defined positive scalar parameter that controls the regularity to impose on the model, leading to a trade-off between data fidelity and regularization. 

The weight $\alpha$ modulates between data-fidelity and regularization, constraining the solution space. Yet, subjecting the entire solution, a dense disparity field, to the same regularity fails to address cases where the assumptions do not hold. Suppose one enforces a smoothness constraint to the output disparity field by simply taking the disparity gradient $\nabla d$. This constraint would incorrectly penalize object boundaries (regions of high image gradients) and hence \cite{godard2017unsupervised,heise2013pm} apply an edge-aware term to reduce the effects of regularization on edge regions. Although the edge-awareness term gives a data-driven approach on regularization, it is still static (the same image will always have the same weights) and independent of the performance of the model. Instead, we propose a space- and training-time varying weighting scheme based on the performance of our model measured by reprojection residuals.

\noindent \textbf{Model-driven adaptive weight.} We propose an adaptive weight $\alpha(x)$ that varies in space and training time for every position $x \in \Omega$ of the solution based on the local residual $\rho(x)=|I(x)-\hat{I}(x)|$ and the global residual, represented by the average per-pixel residual, $\sigma = \displaystyle \frac{1}{\frac{1}{|\Omega|}  \displaystyle \sum_{x \in \Omega}|I(x)-\hat{I}(x)|}$:
\vspace{-1.2em}
\begin{align}
\hspace*{-0.5cm}
\label{residual_weight_eqn}
\alpha(x) &= \exp\left(-\frac{c \rho(x)}{\sigma}\right)
\end{align}
$\alpha$ is controlled by the local residual between an image $I$ and its reprojection $\hat{I}$ at each position while taking into account of the global residual $\sigma$, which correlates to the training time step and decreases over time. $c$ is a scale factor for the range of $\alpha$. $\alpha$ is naturally small when residuals are large and tends to 1 as training converges.

\noindent \textbf{Local adaptation.} Consider a pair of poorly matched  pixels, $(I(x), \hat{I}(x))$, where the residual $|I(x)-\hat{I}(x)|$ is large. By reducing the regularity  on the solution $d(x)$, we effectively allow for exploration in the solution space to find a better match and hence a $d(x)$ that minimizes the data fidelity terms. Alternatively, consider a pair of perfectly matched pixels, $(I(x), \hat{I}(x))$, where $|I(x)-\hat{I}(x)|=0$. We should apply regularization to decrease the scope of the solution space such that we can allow for convergence and propagate the solution. Hence, a spatially adaptive $\alpha(x)$ must vary inversely to the local residual $\rho(x)$ such that we impose regularization when the residual is small and reduce it when the residual is large.

\noindent \textbf{Global adaptation.} Consider a solution $d(x)$ proposed at the first training time step $t=1$. Imposing regularity effectively reduces the solution space based on an assumption about $d(x)$ and biases the final solution. We propose that a weighting scheme $\alpha(x) \rightarrow 1$ as $t \rightarrow \infty$. However, if $\alpha(x)$ is directly dependent on the $t$, then $\alpha(x)$ will change if we continue to train even after convergence -- causing the model to be unstable. Instead, let $\alpha(x)$ be inversely proportional to the global residual $\sigma$ such that $\alpha(x)$ is small when the $\sigma$ is large (generally corresponding to early time steps) and $\alpha(x) \rightarrow 1$ as $\sigma \rightarrow 0$. When training converges (i.e. the global residual has stabilized), $\alpha(x)$ likewise will be stable. This naturally lends to an annealing schedule where $\alpha(x) \rightarrow 1$ as time progresses in training steps. 

\vspace{-0.0em} 
\subsection{Adaptive Regularization}
\vspace{-0.3em}
\label{adaptive_regularization}
Our regularizers assume local smoothness and consistency between the left and right disparities estimated.  We propose to minimize the disparity gradient (smoothness) and the disparity reprojection error (bilateral cyclic consistency) while adaptively weighting both with $\alpha$ (\secref{residual_weights}).

\noindent \textbf{Smoothness loss.} We encourage the predicted disparities to be locally smooth by applying an $L_1$ penalty to the disparity gradients in the x ($\partial_X$) and y ($\partial_Y$) directions. However, such an assumption does not hold at object boundaries, which generally correspond to regions of high changes in pixel intensities; hence, we include an edge-aware term $\lambda$ to allow for discontinuities in the disparity gradient. We also weigh this term adaptively with $\alpha$:
\begin{equation} 
\label{smoothness_loss_eqn}
  	l_{sm} = \sum_{s \in S} \sum_{x \in \Omega} 
  		\alpha^s(x)\lambda^s(x)(|\partial_{X}d^s(x)|+|\partial_{Y}d^s(x)|)
\end{equation}
where $\lambda^s(x) = \exp({-|\nabla^2 I^s(x)|})$ and the $\nabla^2$ operator denotes the image Laplacian. We 
use the image Laplacian over the first order image gradients because it allows the disparity gradients to be aware of intensity changes in both directions. However, we regularize the disparity field using the disparity gradient so that we can allow for independent movement in each direction. Prior to computing the image Laplacian for $\lambda$, we smooth the image with a Gaussian kernel to reduce noise. 

\noindent \textbf{Bilateral cyclic consistency loss.} A common regularization technique in stereo-vision is to maintain the consistency between the left ($d^0$) and right ($d^1$) disparities by reconstructing each disparity through projecting its counter-part with its disparity shifts:
\begin{equation}
\label{lr_reprojection_eqn}
    \hspace{-0.0em}
  	d^{0p}(x) = d^1(x-d^0(x)) \text{ \ and \ } d^{1p}(x) = d^0(x+d^1(x))
\end{equation}
However, in doing so, the projected disparities suffer from the unresolved correspondences of both the disparity ramps, occlusions and dis-occlusions.  
We, propose a bilateral cyclic consistency check that is designed to specifically reason about occlusions while removing the effects of stereo dis-occlusions. We follow the intuition that the disparities $d$ should have an identity mapping when projected to the domain of its stereo-counterpart and back-projected to the original domain as a reconstruction $\hat{d}$ so reconstruction of dis-occlusion is ignored. 
\begin{equation}
\label{bc_reprojection_eqn}
\hspace{-0.2em}
  	\hat{d}^{0}(x) = d^{1p}(x-d^0(x) \text{ \ and \ }  
  	\hat{d}^{1}(x) = d^{0p}(x +d^1(x))
\end{equation}
By applying an $L_1$ penalty on the disparity field and its reconstruction, we are constraining that the cyclic transformations should be the identity transform, which keeps $d^0$ and $d^1$ consistent with each other in co-visible regions. If there exists an occluded region, the region in the reconstruction would be inconsistent with the original -- yielding reprojection error. To avoid penalizing a model for an unresolvable correspondence due to the nature of the data, we propose to adaptively regularize the bilateral cyclic constraint using our residual-based weighting scheme (\eqnref{residual_weight_eqn}). Unsurprisingly, local regions of high reprojection residual often correspond to occluded regions.
\begin{align} 
\label{bc_consistency_loss_eqn} 
  	l_{bc} = \sum_{s \in S} \sum_{x \in \Omega} \alpha^s(x)|d^s(x)-\hat{d}^s(s)|
\end{align}

\vspace{-0.0em}
\section{A Two-Branch Decoder}
\label{two_branch_decoder}
\begin{figure}
\begin{center}
\includegraphics[width=0.35\textwidth]{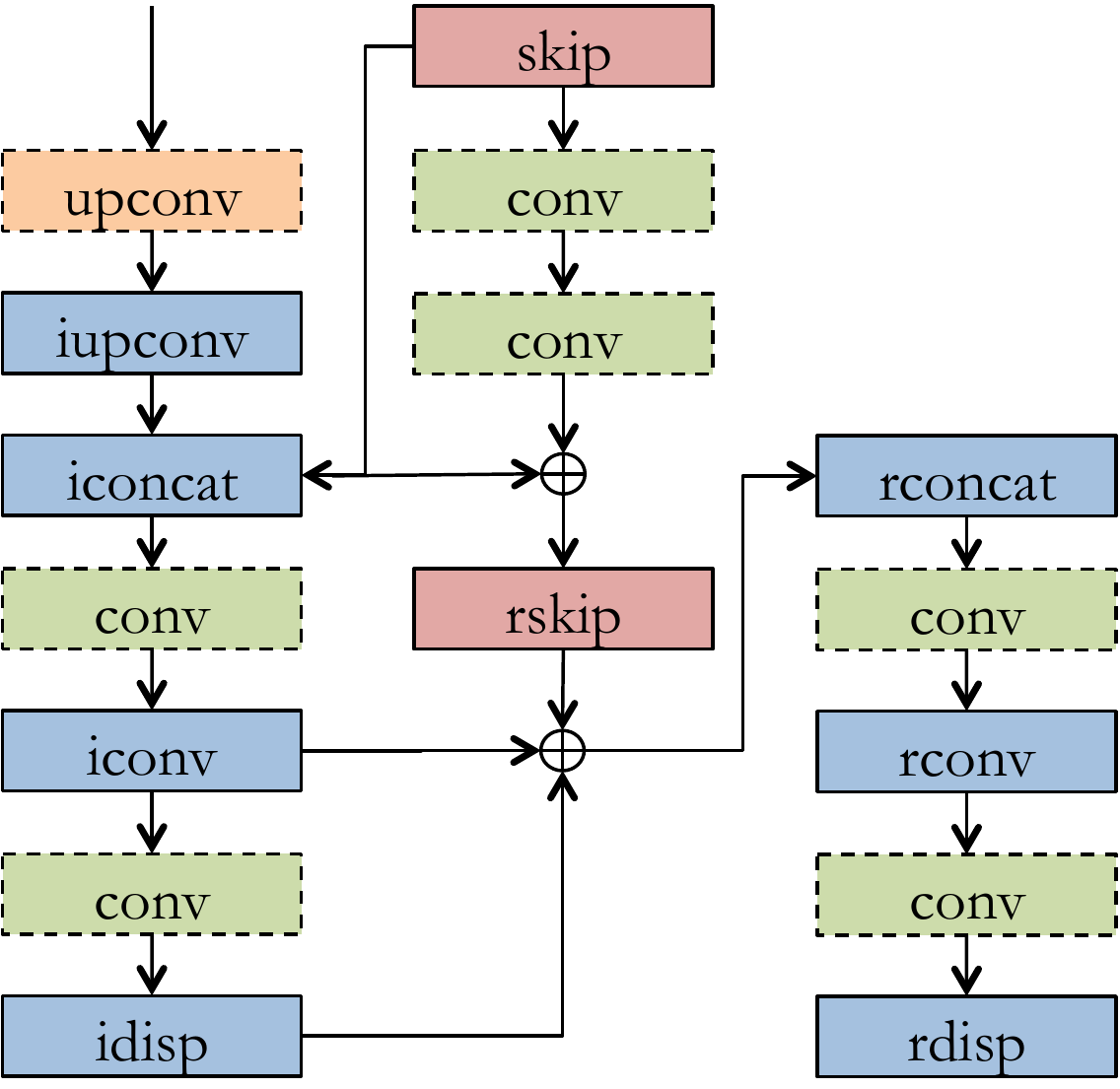}
\vspace{-0.5em}
\caption{Two-branch decoder. \texttt{idisp} produces an initial prediction based only on the data terms and \texttt{rdisp} produces a refined prediction using the entire loss function (\eqnref{loss_function_eqn}). By minimizing just the data terms (\eqnref{loss_function_init_eqn}) in \texttt{idisp}, we force \texttt{iconv} to learn sufficient information for the reconstruction task such that \texttt{rdisp} can utilize such features along with the residual learned from the skip connection to refine a prediction that satisfies data fidelity by imposing regularity based on the data fidelity residual.}
\label{architecture_fig}
\end{center}
\vspace{-1em}
\end{figure}

\begin{figure*}
\centering
\includegraphics[width=0.99\textwidth]{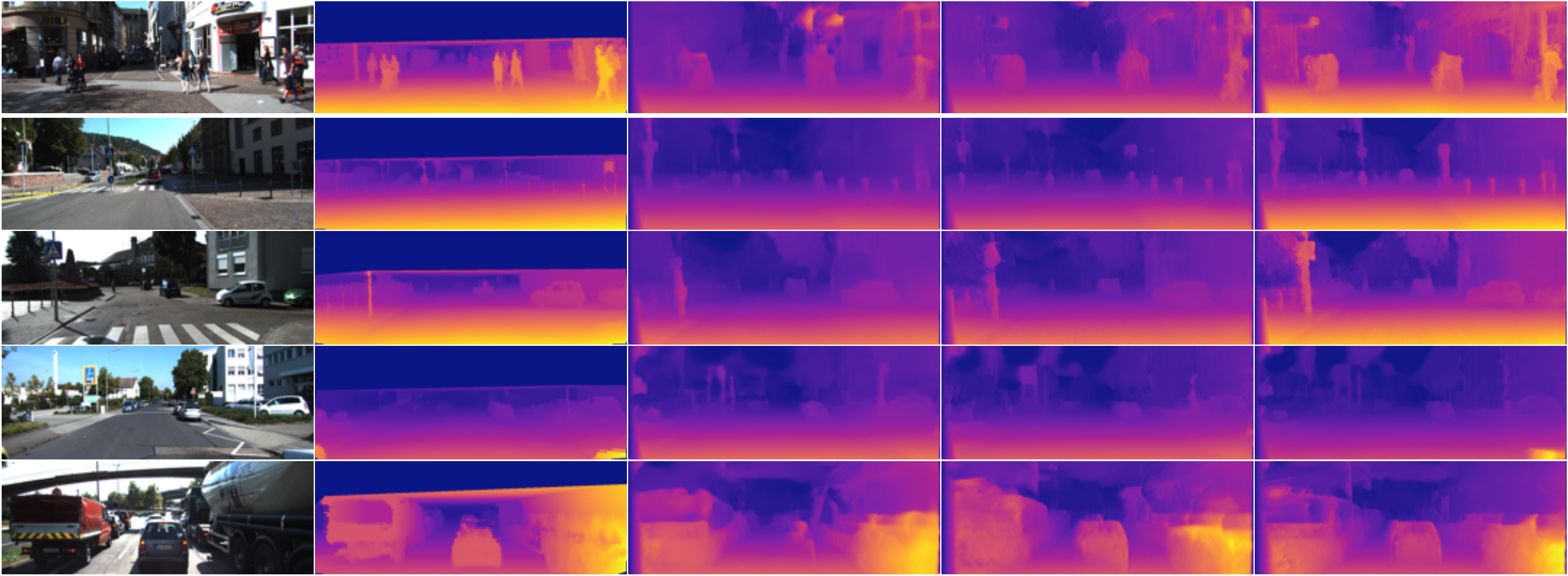}
\vspace{-0.5em}
\caption{Qualitative results on KITTI Eigen split. From left to right: input images, ground-truth disparities, results of Godard et al. \cite{godard2017unsupervised}, our results with a generic decoder and our results with the proposed decoder. Our method under both decoders recovers more scene structures (row 2, 3: street signs, row 5: car in middle). Moreover, the predictions of the proposed two-branch structure are more realistic (row 1: pedestrian on right, row 4: tail of another car at bottom right corner, row 5: hollow trunk of truck on left, where both \cite{godard2017unsupervised} and the generic decoder predicted as a surface).}
\label{eigen_results_fig}
\vspace{-1.5em}
\end{figure*}

\vspace{-0.5em}
As our adaptive weighting scheme (\secref{residual_weights}) is function of the data fidelity residuals, we seek to ensure that the network learns a sufficient representation to minimize the data fidelity loss (\secref{data_unaries}). We propose a two-branch decoder (\figref{architecture_fig}) with one branch (prefixed with `\texttt{i}`) dedicated to learning the features, \texttt{iconv}, necessary to make a prediction that minimizes data fidelity loss:
\begin{equation}
\begin{aligned}  
\label{loss_function_init_eqn}
  	L_{0} = w_{ph}l_{ph}+w_{st}l_{st}
\end{aligned}
\end{equation}
using the reconstructed features via up-convolution and the corresponding \texttt{skip} connection from the encoder. We use a residual block \cite{he2016deep} to learn the skip connection residual, \texttt{rskip}, necessary to minimize \eqnref{loss_function_eqn} -- both data fidelity and regularity loss. By concatenating \texttt{iconv} and \texttt{rskip} with the initial prediction (\texttt{idisp}) as features for the second branch (prefixed with `\texttt{r}`), we have provided the decoder branch with a prediction that satisfies data fidelity along with features necessary to impose regularity. The branch can now utilize such information to refine the initial prediction by adaptively applying regularization based on the data fidelity residual. To maintain a similar network size and run-time, we reduce the depth of the network by 1 and added a single convolution as the first layer to enable a skip connection to the last layer. This, in fact, resulted in our network having $\approx 10$ million fewer parameters than \cite{godard2017unsupervised}. We show qualitative results in \figref{eigen_results_fig} and \ref {kitti_results_fig} where we observe the benefits of learning the features that satisfy data fidelity as we recover more details about the scene geometry. Quantitatively, we show in \tabref{results_eigen_split} and \ref{results_kitti_split} that this structure improves over the state-of-the-art performance on all metrics achieved by our generic encoder with a single branch decoder, where the final predictions of both decoders minimize our objective function (\eqnref{loss_function_eqn}). 

\vspace{-0.0em}
\section{Implementation Details}
\vspace{-0.4em}
\label{implementation_details}
Our approach was implemented using TensorFlow \cite{abadi2016tensorflow}. There are $\approx 31$ million trainable parameters in the generic encoder-decoder \cite{godard2017unsupervised} and $\approx 21$ million in our proposed structure (more details can be found in Supp. Mat. Table 2 and 3). Training takes  $
\approx$ 18 hours using an Nvidia GTX 1080Ti. Inference takes $\approx$ 32 ms per image. We used Adam \cite{kingma2014adam} to optimize our network with a base learning rate of $1.8 \times 10^{-4}$, $\beta_1=0.9$, $\beta_2=0.999$. We then increase the learning rate to $2 \times 10^{-4}$ after 1 epoch, decrease it by half after 46 epochs and by a quarter after 48 epochs for a total of 50 epochs. We use a batch size of 8 with a $512 \times 256$ resolution and 4 levels in our loss pyramid. We are able to achieve our results using the following set of weights for each term in our loss function: $w_{ph}=0.15$, $w_{st}=0.425$, $w_{sm}=0.10$ and $w_{bc}=1.05$. We choose the scale factor $c=5.0$ for the adaptive weight $\alpha$. For our smoothness term, we decrease it by a factor of $2^r$ for each $r$-th resolution in the loss pyramid where $r=0$ refers to our highest resolution at $512 \times 256$ and $r=3$ the lowest. 

Data augmentation is performed online during training. We perform a horizontal flip (with a swap to maintain correct relative positions) on the stereo pairs with $50\%$ probability. Color augmentations on brightness, gamma and color shifts of each channel also occur with $50\%$ chance. We uniformly sample from $[0.5, 1.5]$ for brightness, and $[0.8, 1.2]$ for gamma and each color channel separately.

\begin{table}[t!]
    \setlength\tabcolsep{12pt}
    \scriptsize
    \centering
    \begin{tabular}{l|l}
        \hline
        Metric & Definition \\
        \hline
        AbsRel & $\frac{1}{|\Omega|} \displaystyle \sum_{x \in \Omega} \frac{|z(x)-z_\text{gt}(x)|}{z_\text{gt}(x)}$\\
    
        SqRel & $\frac{1}{|\Omega|} \displaystyle \sum_{x \in \Omega}\frac{|z(x)- z_\text{gt}(x)|^2}{z_\text{gt}(x)}$\\
    
        RMS & $\sqrt{\frac{1}{|\Omega|} \displaystyle \sum_{x \in \Omega}|z(x) - z_\text{gt}(x) |^2}$\\
    
        logRMS & $\sqrt{\frac{1}{|\Omega|} \displaystyle \sum_{x \in \Omega}|\log z(x) - \log z_\text{gt}(x)|^2}$\\
    
        $\log_{10}$ & $\frac{1}{|\Omega|} \displaystyle \sum_{x \in \Omega}|\log z(x)-\log z_\text{gt}(x)|$\\
    
        Accuracy & \% of $z(x)$ s.t. $\delta \doteq \max\big(\frac{z(x)}{z_\text{gt}(x)}, \frac{z_\text{gt}(x)}{z(x)}\big) < \text{threshold}$ \\
        \hline
    \end{tabular}
    \vspace{-0.5em}
    \caption{Error and accuracy metrics. $z(x)$ is the predicted depth at $x \in \Omega$ and $z_\text{gt}(x)$ is the corresponding ground truth. Three different thresholds ($1.25, 1.25^2$ and $1.25^3$) are used in the accuracy metric as a convention in the literature.}
    \label{eval_metrics}
    \vspace{-2.0em}
\end{table}

\begin{table*}[]
\begin{adjustwidth}{-.8in}{-.8in} 
\scriptsize
\centering
\setlength\tabcolsep{10pt}
\begin{tabular}{l c c c c c c c c c}
    \toprule
    & & & \multicolumn{4}{c}{Error Metrics} & \multicolumn{3}{c}{Accuracy Metrics} \\
    \cmidrule(lr){4-7} 
    \cmidrule(lr){8-10}
    Method & Dataset & Cap & AbsRel & SqRel & RMS & logRMS  & $\delta<1.25$ & $\delta<1.25^2$ & $\delta<1.25^3$ \\ \midrule
    Zhou et al. \cite{zhou2017unsupervised}              
    & K  & 80m & 0.208 & 1.768 & 6.856 & 0.283 & 0.678 & 0.885 & 0.957  \\ \midrule
    Mahjourian et al. \cite{mahjourian2018unsupervised} 
    & K  & 80m & 0.163 & 1.240 & 6.220 & 0.250 & 0.762 & 0.916 & 0.968 \\ \midrule
    Garg et al. \cite{garg2016unsupervised}
    & K  & 80m & 0.152 & 1.226 & 5.849 & 0.246 & 0.784 & 0.921 & 0.967 \\ \midrule
    Godard et al. \cite{godard2017unsupervised}            
    & K  & 80m & 0.148 & 1.344 & 5.927 & 0.247 & 0.803 & 0.922 & 0.964  \\ \midrule
    Zhan et al. \cite{zhan2018unsupervised} (w/ video) 
    & K  & 80m & 0.144 & 1.391 & 5.869 & 0.241 & 0.803 & 0.928 & \textbf{0.969}  \\ \midrule
    Ours (Full Model)                      
    & K  & 80m & 0.135 & 1.157 & 5.556 & 0.234 & 0.820 & 0.932 & 0.968  \\ \midrule 
    Ours (Full Model)*                      
    & K  & 80m & \textbf{0.133} & \textbf{1.126} & \textbf{5.515} & \textbf{0.231} & \textbf{0.826} & \textbf{0.934} & \textbf{0.969}  \\ \midrule
    Zhou et al. \cite{zhou2017unsupervised} 
    & CS+K & 80m & 0.198 & 1.836 & 6.565 & 0.275 & 0.718 & 0.901 & 0.960 \\ \midrule
    Mahjourian et al. \cite{mahjourian2018unsupervised} 
    & CS+K & 80m & 0.159 & 1.231 & 5.912 & 0.243 & 0.784 & 0.923 & 0.970 \\ \midrule
    Godard et al. \cite{godard2017unsupervised}                    
    & CS+K  & 80m & 0.124 & 1.076 & 5.311 & 0.219 & 0.847 & 0.942 & 0.973  \\ \midrule
    Ours (Full Model)*                      
    & CS+K  & 80m & \textbf{0.118} & \textbf{0.996} & \textbf{5.134} & \textbf{0.215} & \textbf{0.849} & \textbf{0.945} & \textbf{0.975}  \\ 
    \midrule \midrule
    Zhou et al. \cite{zhou2017unsupervised}             
    & K & 50m & 0.201 & 1.391 & 5.181 & 0.264 & 0.696 & 0.900 & 0.966  \\ \midrule
    Garg et al. \cite{garg2016unsupervised}           
    & K & 50m & 0.169 & 1.080 & 5.104 & 0.273 & 0.740 & 0.904 & 0.962  \\ \midrule
    Godard et al. \cite{godard2017unsupervised}          
    & K & 50m & 0.140 & 0.976 & 4.471 & 0.232 & 0.818 & 0.931 & 0.969  \\ \midrule
    Zhan et al. \cite{zhan2018unsupervised} (w/ video)  
    & K & 50m & 0.135 & 0.905 & 4.366 & 0.225 & 0.818 & 0.937 & \textbf{0.973}  \\ \midrule
    Ours (Full Model)                   
    & K & 50m & 0.128 & 0.856 & 4.201 & 0.220 & 0.835 & 0.939 & 0.972  \\ \midrule
    Ours (Full Model)*          
    & K & 50m & \textbf{0.126} & \textbf{0.832} & \textbf{4.172} & \textbf{0.217} & \textbf{0.840} & \textbf{0.941} & \textbf{0.973} \\ \bottomrule
\end{tabular}
\end{adjustwidth}
\setlength{\belowcaptionskip}{-5pt}
\vspace{-0.5em}
\caption{Quantitative results$^{\ref{eval_metrics}}$ on the KITTI \cite{geiger2012we} Eigen split \cite{eigen2014depth} benchmark. Depths are capped at 50 and 80 meters. K denotes training on KITTI. CS+K denotes pretraining on Cityscape \cite{cordts2016cityscapes} and fine-tuning on KITTI. Our full model using a generic encoder-decoder consistently outperforms other methods in all metrics across both depth caps with the exception of $\delta < 1.25^3$ where \cite{zhan2018unsupervised}, which used temporal information (sequences of stereo-pairs), marginally beats our us by 0.1\%. Our proposed decoder (*) improves over our encoder-decoder model across all metrics and is the state-of-the-art.}
\label{results_eigen_split}
\end{table*}

\vspace{-0.65em}
\section{Experiments and Results}
\vspace{-0.4em}
\label{experiments_results}
We present our results on the KITTI dataset \cite{geiger2012we} under two different training and testing schemes, the KITTI 2015 split \cite{godard2017unsupervised} and the KITTI Eigen split \cite{eigen2014depth,garg2016unsupervised}. The KITTI dataset contains 42,382 rectified stereo pairs from 61 scenes with approximate resolutions of $1242 \times 375$. We evaluate our method on the monocular depth estimation task on KITTI Eigen split and compare our approach with similar variants on a disparity error metric as an ablation study using the KITTI 2015 split. We show that our method outperforms state-of-the-art unsupervised monocular approaches and even supervised approaches on KITTI benchmarks, while generalizing to Make3d \cite{saxena2009make3d}.
\begin{table*}[]
\begin{adjustwidth}{-.9in}{-.9in} 
\scriptsize
\centering
\begin{tabular}{l c c c c c c c c}
\toprule
 & \multicolumn{5}{c}{Error Metrics} & \multicolumn{3}{c}{Accuracy Metrics} \\ 
\cmidrule(lr){2-6} 
\cmidrule(lr){7-9}
Method & AbsRel & SqRel & RMS & logRMS & D1-all & $\delta<1.25$ & $\delta<1.25^2$ & $\delta<1.25^3$ \\ \midrule

\cite{godard2017unsupervised}   w/ Deep3D \cite{xie2016deep3d}                   
& 0.412 & 16.37 & 13.693 & 0.512 & 66.850 & 0.690 & 0.833 & 0.891 \\ \midrule
\cite{godard2017unsupervised}   w/ Deep3Ds \cite{xie2016deep3d}  
& 0.151 & 1.312 & 6.344 & 0.239 & 59.640 & 0.781 & 0.931 & 0.976 \\ \midrule
$ph + st + \lambda^{G} sm$ (\cite{godard2017unsupervised}   w/o Left-Right Consistency)           
& 0.123 & 1.417 & 6.315 & 0.220 & 30.318 & 0.841 & 0.937 & 0.973 \\ \midrule
$ph + st + \lambda^{G} sm + lr$ \cite{godard2017unsupervised}   
& 0.124 & 1.388 & 6.125 & 0.217 & 30.272 & 0.841 & 0.936 & 0.975 \\ \midrule
$ph + st + \alpha \lambda^{G} sm + \alpha lr$ (\cite{godard2017unsupervised} w/ Our Adaptive Regularization)       
& 0.120 & 1.367 & 6.013 & 0.211 & 30.132 & 0.849 & 0.942 & 0.975 \\ \midrule
Aleotti et al. \cite{aleotti2018generative}
& 0.119 & 1.239 & 5.998 & 0.212 & 29.864 & 0.846 & 0.940 & 0.976 \\ \midrule
$ph + st + \lambda^{L} sm + bc$ (Ours w/o Adaptive Regularization)               
& 0.117 & 1.264 & 5.874 & 0.207 & 29.793 & 0.851 & 0.944 & 0.977 \\ \midrule
$ph + st + \alpha \lambda^{L} sm + \alpha lr$ (Ours w/o Bilateral Cyclic Consistency)                
& 0.117 & 1.251 & 5.876 & 0.206 & 29.536 & 0.851 & 0.944 & 0.977 \\ \midrule
$ph + st + \alpha \lambda^{G} sm + \alpha bc$  (Ours w/o Bidirectional Edge-Awareness)
& 0.115 & 1.211 & 5.743 & 0.203 & 28.942 & 0.852 & 0.945 & 0.977 \\ \midrule
$ph + st + \alpha \lambda^{L} sm + \alpha bc$  (Ours Full Model)
& 0.114 & 1.172 & 5.651 & 0.202 & 28.142 & 0.855 & \textbf{0.947} & 0.979 \\ \midrule
$ph + st + \alpha \lambda^{L} sm + \alpha bc$ *  (Ours Full Model w/ 2 Branch Decoder)                            
& \textbf{0.110} & \textbf{1.119} & \textbf{5.576} & \textbf{0.200} & \textbf{27.149} & \textbf{0.856} & \textbf{0.947} & \textbf{0.980} \\ \bottomrule
\end{tabular}
\end{adjustwidth}
\vspace{-0.5em}
\caption{Quantitative comparison$^{\ref{eval_metrics}}$ amongst variants of our model on KITTI 2015 split proposed by \cite{godard2017unsupervised}. Each variant is named according to its loss function. $ph$ and $st$ denote data terms, $sm$ local smoothness, $\alpha$ our adaptive weights, $\lambda^{G}$ image gradients \cite{godard2017unsupervised}, $\lambda^{L}$ image Laplacian, $lr$ left-right consistency \cite{godard2017unsupervised}, and $bc$ our bilateral cyclic consistency. We show the effectiveness of our adaptive regularization (\secref{adaptive_regularization}) by applying it to \cite{godard2017unsupervised} and improving their model. Our full model using a generic encoder-decoder outperforms all variants on every metric, including \cite{aleotti2018generative} which predicts disparities that generate photo-realistic images. Our full model using our proposed two-branch decoder (*) further improves the state-of-the-art.}
\label{results_kitti_split}
\vspace{-0.5em}
\end{table*}

\begin{figure*}
\centering
\includegraphics[width=0.99\textwidth]{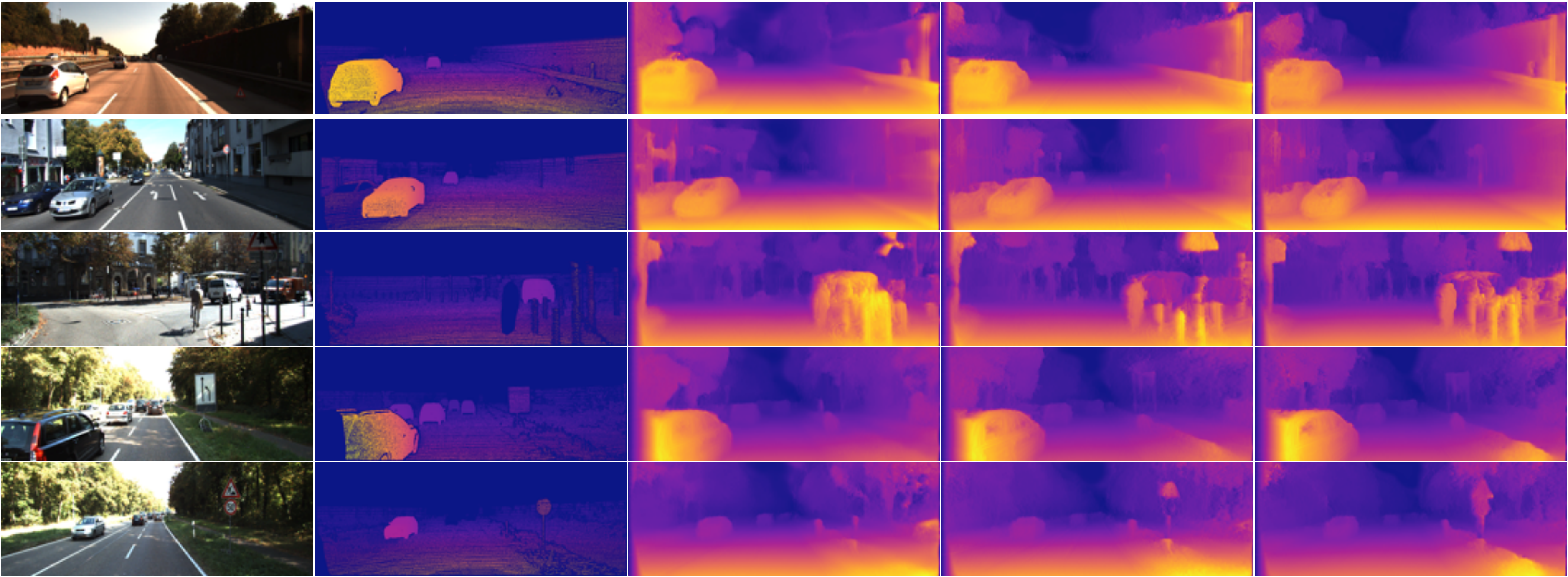}
\vspace{-0.5em}
\caption{Qualitative results on KITTI 2015 split. From left to right: input images, ground-truth depths, results of Godard et al.\cite{godard2017unsupervised}, our results using a generic decoder and our results the proposed decoder. Our approach generates more consistent depths (row 1: walls on right, row 2: building on left) and recovers more detailed structures (row 3: biker and poles on right, rows 4, 5: street signs), with the two-branch decoder recovering the most.}
\label{kitti_results_fig}
\vspace{-1.5em}
\end{figure*}

\begin{figure*}[t!]
    \centering
    \scriptsize
    \subfloat[a][]{
    \raisebox{-.50\height}{\includegraphics[width=0.49\textwidth, height=38mm]{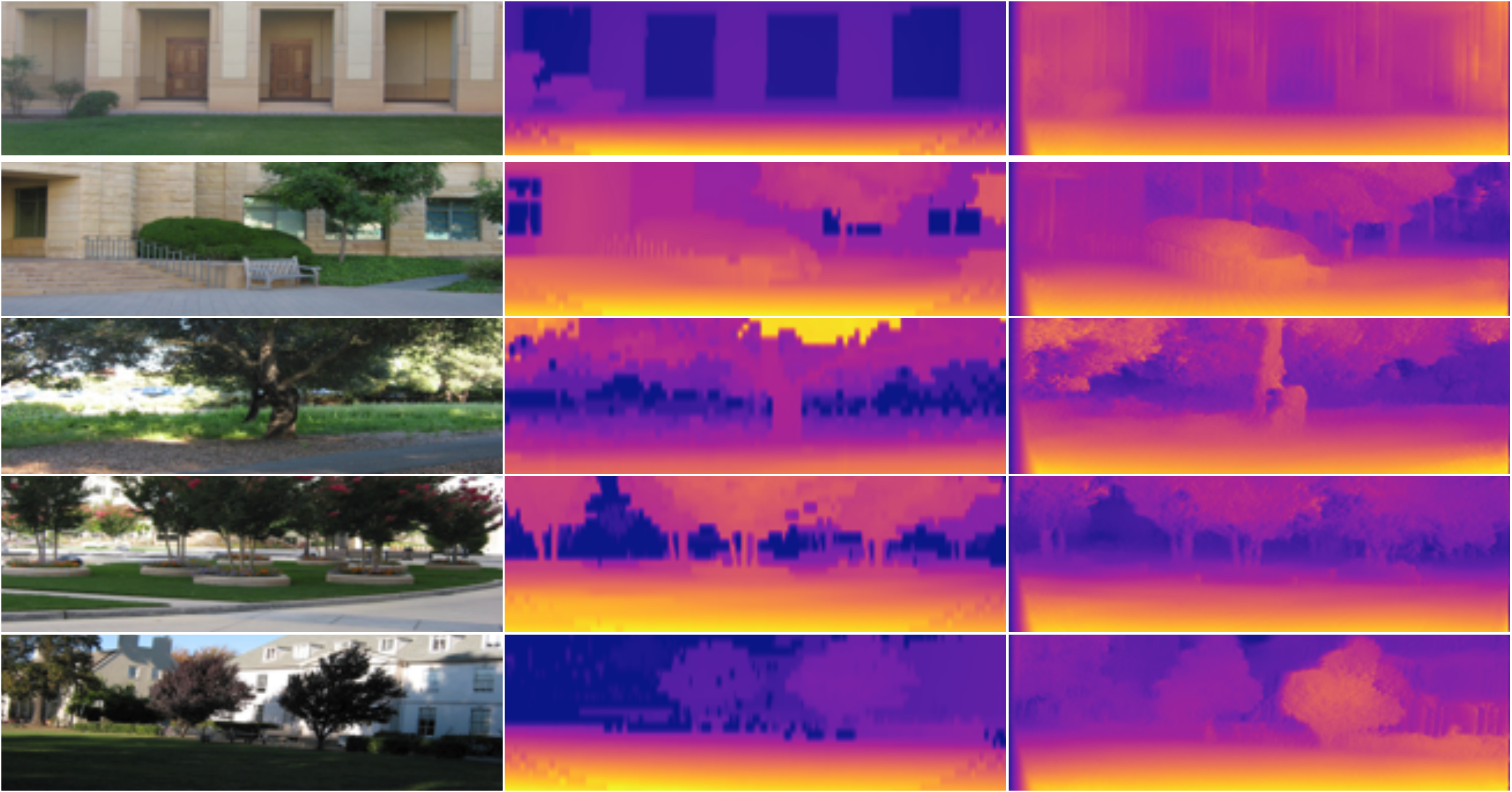}}
    }
    \subfloat[b][]{
    \begin{tabular}{l c c c c c }
        \toprule
        & & \multicolumn{4}{c}{Error Metrics} \\
        \cmidrule(lr){3-6} 
        Method  & Supervised  & AbsRel  &  SqRel  & RMS  & $\text{log}_{10}$  \\ \midrule
        Karsch et al. ~\cite{karsch2012depth} 
        & Yes & 0.417  & 4.894  & 8.172    & 0.144  \\  \midrule         
        Liu et al. ~\cite{liu2016learning} 
        & Yes & 0.462  & 6.625  & 9.972    & 0.161  \\  \midrule
        Laina et al. ~\cite{laina2016deeper} 
        & Yes & {\bf 0.198}  & {\bf 1.665}  & {\bf 5.461}  & {\bf 0.082}  \\  \midrule \midrule
        Godard et al. ~\cite{godard2017unsupervised} 
        & No  & 0.468  & 9.236  & 12.525 & 0.165 \\  \midrule
        Ours 
        & No  & 0.454  & 8.470 & 12.211 & 0.163 \\ \midrule  
        Ours*
        & No  & {\bf 0.427}  & {\bf 8.183} & {\bf 11.781} & {\bf 0.156} \\ \bottomrule
    \end{tabular}
    }
\vspace{-1.5em}
\caption{Qualitative (a) and quantitative (b) results$^{\ref{eval_metrics}}$ on Make3d \cite{saxena2009make3d} with maximum depth of 70 meters. In (a), top to bottom: input images, ground-truth disparities, our results. In (b), unsupervised methods listed are all trained on KITTI Eigen split. Despite being trained on KITTI, we perform comparably to a number of supervised methods trained on Make3d.}
\label{make3d_results_fig}
\vspace{-1.5em}
\end{figure*}

\vspace{-0.0em}
\subsection{KITTI Eigen Split}
\vspace{-0.3em}
We evaluate our method using the KITTI Eigen split \cite{eigen2014depth}, which has 697 test images from 29 scenes. The remaining 32 scenes contain 23,488 stereo pairs, of which 22,600 pairs are used for training and the rest for validation, following \cite{garg2016unsupervised}. We project the velodyne points into the left input color camera frame to generate ground-truth depths. The ground-truth depth maps are sparse ($\approx 5\%$ of the entire image) and prone to errors from rotation of the velodyne and motion of the vehicle and surrounding objects along with occlusions. As a result, we use the cropping scheme proposed by \cite{garg2016unsupervised}, which contains approximately $58\%$ in height and $93\%$ in width of the image dimensions. 

We compare our approach with the recent monocular depth estimation methods at 80 and 50 meters caps in \tabref{results_eigen_split}. \figref{eigen_results_fig} provides a qualitative comparison between our method and the baseline. We note that \cite{zhan2018unsupervised} trained two networks using stereo video streams (as opposed to a single network with stereo pairs like ours and \cite{godard2017unsupervised}), which allows their networks to learn a depth prior in both spatial and temporal domains. Using the network of \cite{godard2017unsupervised} (generic encoder with a single branch decoder), we outperforms all competing methods in all metrics under both depth caps except for $\delta<1.25^3$ where we are comparable to \cite{zhan2018unsupervised}. We improve consistently over \cite{godard2017unsupervised} and \cite{zhan2018unsupervised}
by an average of 8.7\% and 5.75\% in AbsRel, 13.1\% and 10.5\% in SqRel and even 5.25\% and 2.55\% in logRMS, respectively. Furthermore, we score significantly higher in $\delta<1.25$ (the hardest accuracy metric), which suggests that our model produces more correct and realistically detailed depths than all competing methods. In addition, our two-branch decoder improves over the said results across all metrics and depth caps and is the current state-of-the-art. \tabref{results_eigen_split} shows that our model also beats \cite{godard2017unsupervised} when pretraining on Cityscape \cite{cordts2016cityscapes} and fine-tuning on KITTI. An ablation study on Eigen Split examining the effects of each of our contributions (\secref{adaptive_regularization}) can be found in our Supp. Mat.

\vspace{-0.0em}
\subsection{KITTI 2015 Split}
\vspace{-0.3em}
We evaluate our method on 200 high quality disparity maps provided as part of the official KITTI training set \cite{geiger2012we}. These 200 stereo pairs cover 28 of the total 61 scenes. From 30,159 stereo pairs covering the remaining 33 scenes, we choose 29,000 for training and the rest for validation. While typical training and evaluation schemes project velodyne laser values to depth, we choose to use the provided disparity maps as they are less erroneous than velodyne data points. In addition, we also use the official KITTI disparity metric of end-point-error (D1-all) to measure our performance as it is a more appropriate metric on our class of approach that outputs disparity and synthesizes depth from the output using camera focal length and baseline.

We show qualitative comparisons in \figref{kitti_results_fig} and quantitative comparisons in \tabref{results_kitti_split}. \tabref{results_kitti_split} also serves as an ablation study on variants belonging to the stereo unsupervised paradigm using different image formation model and regularization terms. We show that by simply applying our adaptive regularization to \cite{godard2017unsupervised}, we achieve improvement over their model. We also study the effects of substituting our bilateral cyclic consistency with the left-right consistency regularizer \cite{godard2017unsupervised}. We also substitute image Laplacian with image gradients for edge-aware weights. In addition, we find that adaptive regularization and bilateral cyclic consistency contribute similarly to the improvements of the models. However, when combined they achieve significantly improvements over the baseline method (and all variants) in every metric. Furthermore, when using our proposed decoder, we again surpass all variants on every metric. We additionally outperform \cite{aleotti2018generative}, who uses a GAN to constrain the output disparities to produce photo-realistic images during reconstruction. This result aligns with our performance on accuracy metrics -- our method produces accurate and realistic depths. 

\vspace{-0.0em}
\subsection{Generalizing to Different Datasets: Make3d}
\vspace{-0.4em}
To show that our model generalizes, we present our qualitative and quantitative results in \figref{make3d_results_fig} on the Make3d dataset \cite{saxena2009make3d} containing 134 test images with $2272 \times 1707$ resolution. Make3d provides range maps (resolution of $305 \times 55$) for ground-truth depths, which must be rescaled and interpolated. We use the central cropping proposed by \cite{godard2017unsupervised} where we generate a $852 \times 1707$ crop centered on the image. We use the standard $C1$ evaluation metrics$^{\ref{eval_metrics}}$ proposed for Make3d and limit the maximum depth to 70 meters. The results of the supervised methods are taken from \cite{godard2017unsupervised}. Because Make3d does not provide stereo pairs, we are unable to train on it. However, we find that despite having trained our model on KITTI Eigen split, our performance is comparable to that of supervised methods trained on Make3d and is better than the baseline across all metrics.  

\vspace{-0.0em}
\section{Discussion}
\vspace{-0.4em}
\label{discussion}
In this work, we proposed an adaptive weighting scheme (\secref{adaptive_regularization}) that is both spatially and time varying, allowing for not only a data-driven, but also model-driven approach to regularization. Moreover, we introduce a bilateral cyclic consistency constraint that not only enforces consistency between the left and right disparities, but also removes stereo dis-occlusions while discounting unresolved occlusions when combined with our weighting scheme. Finally, we propose a two-branch decoder that achieves the state-of-the-art by learning features to improve data residual for imposing our adaptive regularity. We achieve state-of-the-art performance on two KITTI benchmarks and show that our method generalizes to Make3d. Our two-branch decoder further improves over those results. Our experiments (\tabref{results_eigen_split} and \ref{results_kitti_split}) show that our approach produces depth maps with more details while maintaining global correctness.
 
In future work, we plan to improve robustness to specular and transparent surfaces as these regions tend to produce inconsistent depths. We are also exploring more sophisticated regularizers in place of the simple disparity gradient. Finally, we believe that the task should drive the network architecture. Rather than using a generic network, finding a better architectural fit could prove to be ground-breaking and further push the state-of-the-art.

\noindent{\bf Acknowledgements.} This work was supported by NRF-2017R1A2B4006023, NRF-2018R1A4A1059731, ONR N00014-17-1-2072, ARO W911NF-17-1-0304.

{\small
\bibliographystyle{ieee}
\bibliography{egbib}
}

\newpage

\onecolumn

\appendix

\section{Problem Formulation}
In this section, we give the formulation for predicting the disparities for a single view using stereo imagery as supervision. Given a single image $I^0$, our goal is to estimate a function $d = f(I^0, \omega) \in \mathbb{R}_{+}$ that represents the disparity of $I^0$, where $f$ is a network parameterized by $\omega$. We assume $I^0$ belongs to a stereo-pair $(I^0, I^1)$ with which we exploit $I^1$ to learn a representation $d$ for predicting scene geometry from $I^0$ by maximizing the posterior distribution:
\begin{align}
\label{posterior_eqn}
	p(d|I^0,I^1) \propto  p(I^1|I^0, d) \cdot p(I^0, d)
\end{align}
We assume that both the likelihood $p(I^1|I^0, d)$ and the prior $p(I^0, d)$ follow a Laplacian distribution. The likelihood can be approximated by a data fidelity term $g(I^0, I^1, d)$ and the prior by a regularization term $h(I^0, d)$ and will have the form:
\begin{align}
\label{likelihood_eqn}
	p(I^1|I^0, d) \approx \exp(-\frac{g(I^0, I^1, d)}{a})
\end{align}
\begin{align}
\label{prior_eqn}
	p(I^0, d) \approx \exp(-\frac{h(I^0, d)}{b})
\end{align}
We then take the negative log to form our data and regularization terms:
\begin{equation}
\label{derivation_eqn}
\begin{aligned}
	-\log p(d|I^0, I^1) &\propto -\log \exp(-\frac{g(I^0, I^1, d)}{a}) \cdot \exp(-\frac{h(I^0, d)}{b}) \\
	&\propto \frac{1}{a} g(I^0, I^1, d) + \frac{1}{b} h(I^0, d) \\
	&\propto \underbrace{g(I^0, I^1, d)}_{\text{data fidelity}} + \alpha \underbrace{h(I^0, d)}_{\text{regularization}}
\end{aligned}
\end{equation}
Given $I^1$ and $d$, we can derive $\hat{I}^0 = I^1(x+d(x))$. We substitute $g(I^0, I^1, d)$ with a generic image reconstruction function:
\begin{equation}
\label{data_fidelity_eqn}
\begin{aligned}
g(I^0, I^1, d) = \sum_{x \in \Omega} | I^0(x)-\hat{I}^0(x) |
\end{aligned}
\end{equation}
We can similarly substitute $h(I^0, d)$ with a generic prior such as local smoothness with edge-awareness:
\begin{equation}
\label{regularization_eqn}
\begin{aligned}
h(I^0, d) = \sum_{x \in \Omega} \lambda^0(x)( |\partial_{X}d^0(x)|+|\partial_{Y}d^0(x)|)
\end{aligned}
\end{equation}

\begin{figure*}[!ht]
\centering
\includegraphics[width=0.95\textwidth]{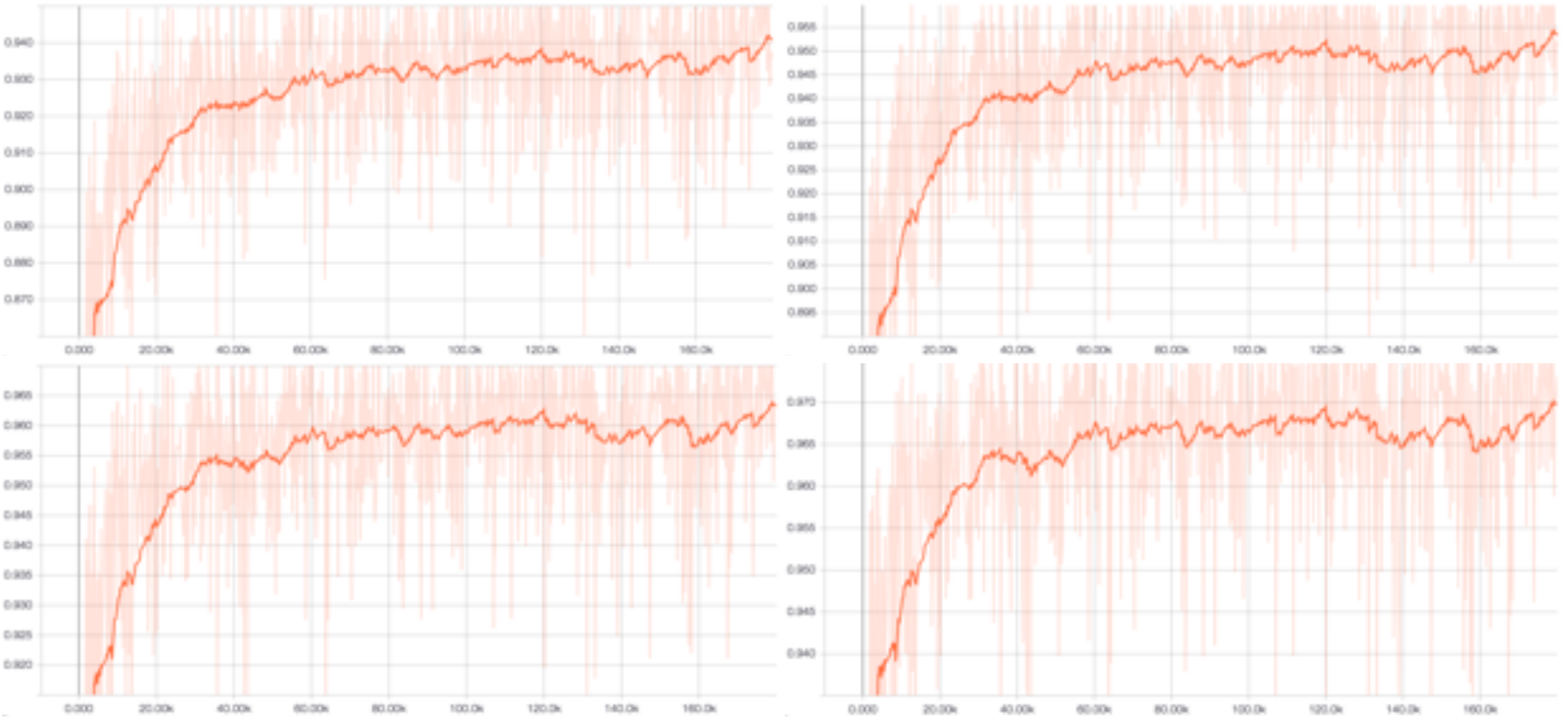}
\caption{An example of the behavior of our residual-based adaptive weighting scheme (as describe in Sec. 3.2 from main text) in the process of training our model. Top row from left to right: the mean of $\alpha$ applied to resolution $r=0$ (full resolution) and resolution $r=1$ (one-half resolution) of the loss pyramid. Bottom row from left to right: mean of $\alpha$ applied to resolution $r=2$ (one-quarter resolution) and resolution $r=3$ (one-eighth resolution). The value recorded (light orange) is the mean of the weights that is being applied to the regularization terms of our model. A weight $\alpha(x)$ is assigned to each $x \in \Omega$ position of the solution; hence, the set of weights $\alpha$ will vary spatially across the image domain. The trend line (dark orange) shows that the average weight for regularization is increasing over time (implying that residual is decreasing over time) -- the training time varying property exhibited by our adaptive weighting scheme.}
\label{adaptive_weights_fig}
\end{figure*}

\newpage

\section{Adaptive Regularization}

Our adaptive regularization weighting scheme $\alpha$ (Sec. 3.2 from main text) allows us to have a model-driven approach to regularization. While a number of literature have exploited a data-driven approach to regularization by weighting the amount of regularity imposed by the structure of the data (e.g. gradients of an image \cite{hoiem2007recovering,godard2017unsupervised}), this class of approaches are still static in terms of weights for a given example as the same image will always give the same weights. Our approach is both model-driven as well as data-driven as our regularization weighting scheme (Eqn. 6 from main text) is a function of the output of the model and the data-fidelity residual (as determined by reconstruction of the images). 

Traditionally, the weight of the regularization terms is a static scalar. However, we argue that imposing the same amount of regularity to the entire solution fails to address corner cases. We propose that $\alpha$ should be spatially varying and inversely proportional to the local residual. While the notion of ``trusting'' the prior (or the regularizers) when data-fidelity fails to explain the scene is intuitive, this assumption is only valid once we are able to sufficiently satisfy the data-fidelity term; otherwise we are restricting our model to a biased set of solutions without having fully explored the solution space. This is apparent in the example given in Sec. 3.2 of the main text regarding the training time varying property of $\alpha$ -- a solution proposed at the first time step cannot be trusted in terms of its data-fidelity and hence we should not impose regularity. Therefore, we propose that our weighting scheme $\alpha \rightarrow 0$ when residual is large and $\alpha \rightarrow 1$ as the residual tends to 0. Naturally the data-fidelity residual decreases over time as the training progresses, which we exploit instead of directly making $\alpha$ a function of training time. Thus, when the model converges, $\alpha$ will also converge.   

We apply our adaptive weighting scheme to each of the regularization terms (Sec. 3.3 from main text) at each level of the loss pyramid for a total of four levels beginning from the full resolution to one-eighth resolution. \figref{adaptive_weights_fig} shows the behavior of the adaptive weights on each level of the loss pyramid. The figure was taken from a training process. The recorded value in dark orange is the trend line representing the mean of the weights that is being applied to our solution. We see that as the model improves our adaptive weights proportionally increase, which equivalently impose regularity on the model. 

\begin{figure*}[ht!]
\begin{center}
\includegraphics[width=1.00\textwidth]{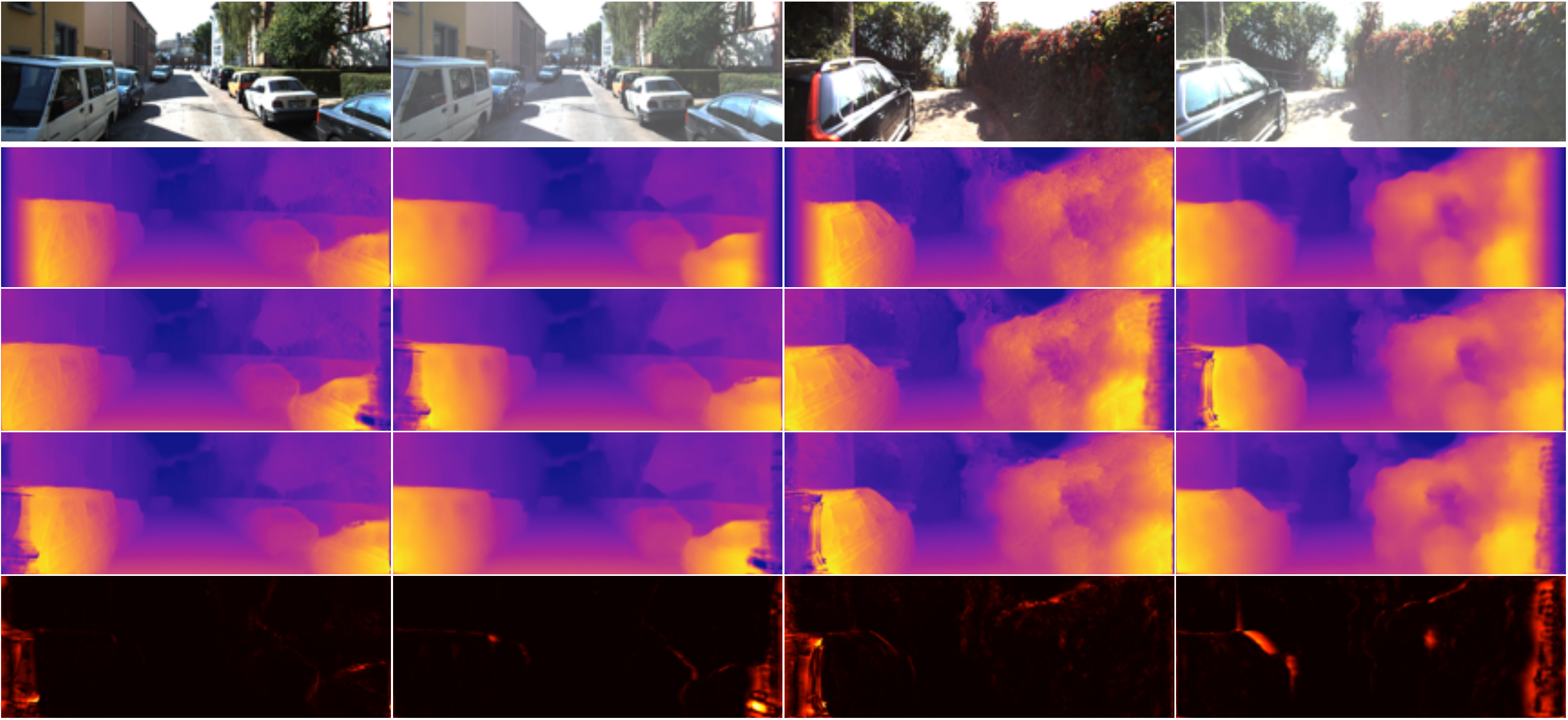}
\end{center}
\setlength{\belowcaptionskip}{-12pt}
\caption{An illustration of our bilateral cyclic constraint with two examples (each example spans two columns). Top to bottom: left image that serves as input to the network with a whitened right image used only in the loss function, initial left and right disparities, projected left and right disparities, reconstructed left and right disparities through back-projection, and absolute difference between the initial disparities and their reconstructions. Our network predicts both the left and right disparities associated with the input left image to enable us to enforce our bilateral cyclic constraint. The right image (whitened) is not given to the network and is only shown to give a frame of reference for the right disparities predicted. From the heat maps (last row), we see that the regions of high intensity (error) is associated with occlusion boundaires (e.g. the missing sections of the cars in the left side of the imagse along with the edges of the cars). In such cases where there are no correspondences (yielding high reconstruction error), our adaptive regularization scheme will discount such regions.}
\label{bc_decoder_fig}
\end{figure*}

\newpage

\section{Bilateral Cyclic Consistency}
In this work, we propose using a bilateral cyclic constraint as a way of regularizing the behavior of the predicted disparities. Our model enforces bilateral cyclic consistency (Sec. 3.3 from main text) by projecting the set of disparities of a given image in a stereo pair to its counter-part and back-projected to itself as a reconstruction (Eqn. 9 from main text). We then apply an $L1$ penalty to the difference between the initial disparities and its reconstruction. By doing so, we are imposing regularity on the disparity map in its original frame of reference as opposed to a relative frame of reference (i.e. left-right consistency). By enforcing the cyclic application of the disparities to be the identity, we constrain that the  correspondences found in both images are co-visible regions. In the case of occlusions, the constraint will be violated, yielding high loss. While a solution would be to discount such errors by setting an arbitrary threshold (as attempting to reconstruct a region that does not exist can never yield a correct solution), our adaptive weighting scheme (Sec. 3.2 from main text) provide an elegant solution to such scenarios. As one can never find correct correspondences for occluded regions of two images, this naturally lends to a high data-fidelity residual with which we adaptively discount using $\alpha$. 

We illustrate the cyclic application of the left and right disparities in \figref{bc_decoder_fig}. Although we are never given the right image (whitened image) of the stereo pair, our network successfully hallucinates its existence and predicts the corresponding disparities to enable our bilateral cyclic consistency check. The areas of inconsistency between the initial disparities and their reconstructions (last row of \figref{bc_decoder_fig}) exist near the occlusion boundaries, which we use $\alpha$ to effectively guide our model to regularize these regions to improve depth consistency.

\begin{table*}[]
\begin{adjustwidth}{-1.2in}{-1.2in} 
\fontsize{8}{10}\selectfont
\centering
\begin{tabular}{l c c c c c c c}
\toprule
& \multicolumn{4}{c}{Error Metrics} & \multicolumn{3}{c}{Accuracy Metrics} \\
\cmidrule(lr){2-5} 
\cmidrule(lr){6-8}
Method & AbsRel & SqRel & RMS & logRMS  & $\delta<1.25$ & $\delta<1.25^2$ & $\delta<1.25^3$ \\ \midrule
& \multicolumn{7}{c}{80 Meter Depth Cap} \\
\midrule
$ph + st + \lambda^{G} sm + lr$ (Godard et al. \cite{godard2017unsupervised})          
& 0.148 & 1.344 & 5.927 & 0.247 & 0.803 & 0.922 & 0.964  \\ \midrule
Zhan et al. \cite{zhan2018unsupervised} (w/ video) 
& 0.144 & 1.391 & 5.869 & 0.241 & 0.803 & 0.928 & \textbf{0.969}  \\ \midrule
$ph + st + \alpha \lambda^{G} sm + \alpha lr$ (\cite{godard2017unsupervised} w/ Our Adaptive Regularization)  
& 0.145 & 1.302 & 5.790 & 0.245 & 0.807 & 0.923 & 0.965  \\ \midrule
$ph + st + \lambda^{L} sm + bc$ (Ours w/o Adaptive Regularization)  
& 0.141 & 1.266 & 5.761 & 0.241 & 0.811 & 0.925 & 0.965  \\ \midrule
$ph + st + \alpha \lambda^{L} sm + \alpha lr$ (Ours w/o Bilateral Cyclic Consistency)
& 0.140 & 1.290 & 5.746 & 0.238 & 0.817 & 0.929 & 0.967  \\ \midrule
$ph + st + \alpha \lambda^{G} sm + \alpha bc$  (Ours w/o Bidirectional Edge-Awareness)
& 0.138 & 1.191 & 5.637 & 0.237 & 0.817 & 0.929 & 0.967  \\ \midrule
$ph + st + \alpha \lambda^{L} sm + \alpha bc$  (Ours Full Model)                
& 0.135 & 1.157 & 5.556 & 0.234 & 0.820 & 0.932 & 0.968  \\
\midrule
Ours (Full Model) \% Improvement over \cite{godard2017unsupervised} 
& {\color{blue} 8.8\%}
& {\color{blue} 13.9\%} 
& {\color{blue} 6.3\%}
& {\color{blue} 5.3\%}
& {\color{blue} 2.1\%} 
& {\color{blue} 1.1\%}
& {\color{blue} 0.4\%} \\ 
\midrule 
Ours (Full Model) \% Improvement over \cite{zhan2018unsupervised}         
& {\color{blue} 6.3\%}
& {\color{blue} 16.8\%} 
& {\color{blue} 5.3\%}
& {\color{blue} 2.9\%}
& {\color{blue} 2.1\%} 
& {\color{blue} 0.4\%}
& {\color{red} 0.1\%} \\ 
\midrule 
$ph + st + \alpha \lambda^{L} sm + \alpha bc$ *  (Ours Full Model w/ 2 Branch Decoder)                       
& \textbf{0.133} & \textbf{1.126} & \textbf{5.515} & \textbf{0.231} & \textbf{0.826} & \textbf{0.934} & \textbf{0.969} \\ \midrule
Ours (Full Model)* \% Improvement over \cite{godard2017unsupervised} 
& {\color{blue} \textbf{10.1\%}}
& {\color{blue} \textbf{16.2\%}} 
& {\color{blue} \textbf{7.0\%}}
& {\color{blue} \textbf{6.5\%}}
& {\color{blue} \textbf{2.9\%}} 
& {\color{blue} \textbf{1.3\%}}
& {\color{blue} \textbf{0.5\%}} \\ 
\midrule 
Ours (Full Model)* \% Improvement over \cite{zhan2018unsupervised}         
& {\color{blue} \textbf{7.6\%}}
& {\color{blue} \textbf{19.1\%}} 
& {\color{blue} \textbf{6.0\%}}
& {\color{blue} \textbf{4.1\%}}
& {\color{blue} \textbf{2.9\%}} 
& {\color{blue} \textbf{0.6\%}}
& {\color{blue} \textbf{0.0\%}} \\ \midrule
& \multicolumn{7}{c}{50 Meter Depth Cap} \\
\midrule

$ph + st + \lambda^{G} sm + lr$ (Godard et al. \cite{godard2017unsupervised})          
& 0.140 & 0.976 & 4.471 & 0.232 & 0.818 & 0.931 & 0.969  \\ \midrule
Zhan et al. \cite{zhan2018unsupervised} (w/ video)  
& 0.135 & 0.905 & 4.366 & 0.225 & 0.818 & 0.937 & \textbf{0.973}  \\ \midrule
$ph + st + \alpha \lambda^{G} sm + \alpha lr$ (\cite{godard2017unsupervised} w/ Our Adaptive Regularization)  
& 0.138 & 0.957 & 4.417 & 0.230 & 0.824 & 0.933 & 0.970  \\ \midrule
$ph + st + \lambda^{L} sm + bc$ (Ours w/o Adaptive Regularization) 
& 0.134 & 0.944 & 4.389 & 0.227 & 0.825 & 0.934 & 0.969  \\ \midrule
$ph + st + \alpha \lambda^{L} sm + \alpha lr$ (Ours w/o Bilateral Cyclic Consistency)  
& 0.133 & 0.942 & 4.351 & 0.224 & 0.832 & 0.937 & 0.971  \\ \midrule
$ph + st + \alpha \lambda^{G} sm + \alpha bc$  (Ours w/o Bidirectional Edge-Awareness)
& 0.131 & 0.881 & 4.265 & 0.224 & 0.832 & 0.937 & 0.971  \\ \midrule
$ph + st + \alpha \lambda^{L} sm + \alpha bc$  (Ours Full Model)              
& 0.128 & 0.856 & 4.201 & 0.220 & 0.835 & 0.939 & 0.972  \\  
\midrule
Ours (Full Model) \% Improvement over \cite{godard2017unsupervised}        
& {\color{blue} 8.6\%}
& {\color{blue} 12.3\%}
& {\color{blue} 6.0\%}
& {\color{blue} 5.2\%} 
& {\color{blue} 2.1\%}
& {\color{blue} 0.9\%}
& {\color{blue} 0.3\%} \\ \midrule
Ours (Full Model) \% Improvement over \cite{zhan2018unsupervised}         
& {\color{blue} 5.2\%}
& {\color{blue} 5.4\%}
& {\color{blue} 3.8\%}
& {\color{blue} 2.2\%} 
& {\color{blue} 2.1\%} 
& {\color{blue} 0.2\%} 
& {\color{red} 0.1\%} \\ \midrule
$ph + st + \alpha \lambda^{L} sm + \alpha bc$ *  (Ours Full Model w/ 2 Branch Decoder)          
& \textbf{0.126} & \textbf{0.832} & \textbf{4.172} & \textbf{0.217} & \textbf{0.840} & \textbf{0.941} & \textbf{0.973} \\ \midrule
Ours (Full Model)* \% Improvement over \cite{godard2017unsupervised}        
& {\color{blue} \textbf{10.0\%}}
& {\color{blue} \textbf{14.7\%}}
& {\color{blue} \textbf{6.7\%}}
& {\color{blue} \textbf{6.5\%}} 
& {\color{blue} \textbf{2.7\%}}
& {\color{blue} \textbf{1.1\%}}
& {\color{blue} \textbf{0.4\%}} \\ \midrule
Ours (Full Model)* \% Improvement over \cite{zhan2018unsupervised}         
& {\color{blue} \textbf{6.7\%}}
& {\color{blue} \textbf{8.1\%}}
& {\color{blue} \textbf{4.4\%}}
& {\color{blue} \textbf{3.6\%}} 
& {\color{blue} \textbf{2.7\%}} 
& {\color{blue} \textbf{0.4\%}} 
& {\color{blue} \textbf{0.0\%}} \\
\bottomrule
\end{tabular}
\end{adjustwidth}
\caption{Ablation study on the KITTI Eigen split \cite{garg2016unsupervised}. We compare each variant of our model to the top performing methods in the monocular depth prediction task. Our full model using a generic single-branch decoder consistently outperforms the best previous methods \cite{godard2017unsupervised, zhan2018unsupervised} in all metrics across both depth caps. Each of our partial models improves over the baseline \cite{godard2017unsupervised} consistently across all metrics and depth caps. In fact, our model without using our bidirectionally informed edge-awareness is already able to exceed the performance of \cite{zhan2018unsupervised} on most metrics, despite not using temporal knowledge and multiple networks as \cite{zhan2018unsupervised} did, with the exception of $\delta<1.25^3$, where \cite{zhan2018unsupervised} marginally beat us by approximately 0.1\%. Our full model using our two-branch decoder (marked by *) outperforms all variants across all metrics and depth caps and is the state-of-the-art. We show the relative percentage boost in performance in all metrics in the two rows following the results of our full model using a single-branch decoder and our full model using a two-branch decoder(*).}
\label{ablation_eigen_split}
\end{table*}

\section{Ablation Studies on Eigen Split}

\tabref{ablation_eigen_split} shows an ablation study on the KITTI Eigen split benchmark. The two rows following our full model and our full modeling using the proposed two-branch decoder (Sec. 4 of main paper) denotes the percentage improvement over  \cite{godard2017unsupervised} and \cite{zhan2018unsupervised}. Our full model using a generic encoder with a single-branch decoder outperforms the top performing methods \cite{godard2017unsupervised, zhan2018unsupervised} across all metrics under both depth caps. Notably, we improve over \cite{godard2017unsupervised} and \cite{zhan2018unsupervised} by an average of 8.7\% and 5.75\% in AbsRel, 13.1\% and 11.1\% in SqRel and even 5.25\% and 2.55\% in logRMS, respectively. Furthermore, we improve 2.1\% over both in $\delta < 1.25$ (the hardest accuracy metric). \cite{zhan2018unsupervised} is only able to outperform our full model marginally in the $\delta<1.25^3$ metric by 0.1\% despite using multiple networks and stereo video streams for training as opposed to stereo pairs. 

Moreover, each of our partial models using the generic encoder with the single-branch decoder (same network as \cite{godard2017unsupervised}) shows improvement over the baseline \cite{godard2017unsupervised}. Even by simply applying our adaptive regularization to \cite{godard2017unsupervised}, we improve consistently across all metrics. More importantly, our model without bidirectionally informed edge-awareness is already able to outperform \cite{zhan2018unsupervised} on most metrics across both depth caps with the exception of $\delta<1.25^2$ and $\delta<1.25^3$ where we are comparable. Based on \tabref{ablation_eigen_split}, we can see that each of our individual contribution improves the model. There is a significant performance gain when multiple contributions are applied to the model (e.g. adaptive regularization with bilateral cyclic consistency versus \cite{godard2017unsupervised} with adaptive regularization). The strongest model is produced when all of the contributions are combined as each contribution complements the others to resolve inconsistencies in object boundaries, co-visible and occluded regions.

Furthermore when applying our full model using the proposed two-branch decoder, we further improve over all methods. Specifically, we improve over \cite{godard2017unsupervised} and \cite{zhan2018unsupervised} by an average of 10.05\% and 7.15\% in AbsRel, 15.45\% and 13.6\% in SqRel, 6.85\% and 5.2\% in RMS, and 6.5\% and 3.85\% in logRMS, respectively. We even improve 2.8\% over both in $\delta < 1.25$. \cite{zhan2018unsupervised} is comparably to us in  $\delta<1.25^3$ metric. Our full model using the proposed two-branch decoder is the state-of-the-art in the unsupervised single image depth prediction task. 

\section{Qualitative Comparison Between Single-Branch and Two-Branch Decoder}
\begin{figure*}[hb]
\begin{center}
\includegraphics[width=1.00\textwidth]{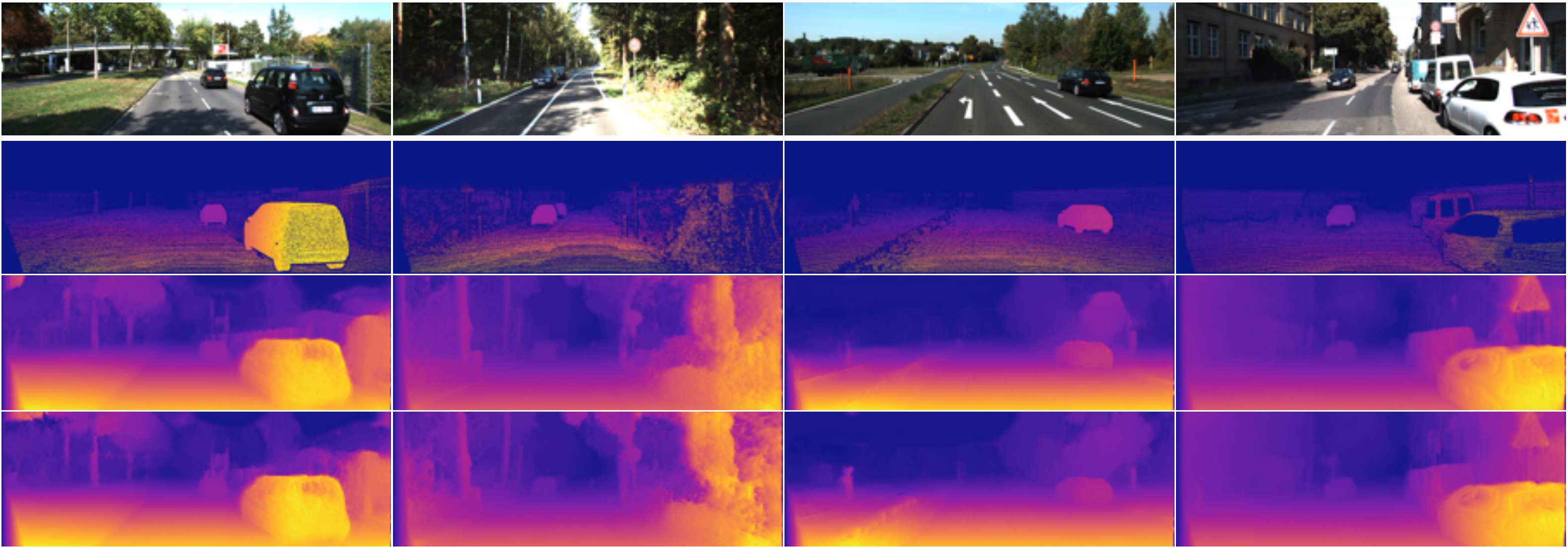}
\end{center}
\setlength{\belowcaptionskip}{-12pt}
\caption{A qualitative comparison between the predictions of a single-branch decoder and the proposed two-branch decoder. Top to bottom: the input image, ground-truth disparities, the results of a single-branch decoder, and the results of the proposed two-branch decoder. The two-branch decoder consistently produces more detailed disparity maps that recovers more of the scene geometry. In the left-most (first) column, while both decoders are able to correct predict the scene globally, the single-branch decoder is unable to recover the details of sign in the distant whereas the two-branch decoder is able to fully recover it. In the second column, the two-branch decoder can recover both of the poles for the sign on the right. In the third column, the two-branch decoder is able to recover the small red pole on the left whereas the single-branch decoder can only recover it partially. In the right-most column, the two-branch decoder is able to recover the small pole on the right next to the wall, the sign in the distance next to the car, the sign above the van and the pole for the sign on the right.}
\label{vs_decoder_fig}
\end{figure*}

It is well-known that geometry can be recovered in co-visible regions (barring texture-less regions) simply by establishing correspondence between two views of the scene. Our proposed decoder (Fig. 1 from main text) dedicates one branch to learning the necessary features to satisfy data-fidelity, which in this case is the reconstruction between the stereo pairs. In doing so, we also produce an initial solution that satisfies data-fidelity. The second branch then aims to refine such a solution by learning the residual features from the skip connection necessary for adaptively imposing regularity. \figref{vs_decoder_fig} gives a qualitative comparison between the single-branch decoder (row 3) and the proposed two-branch decoder (row 4). We see that the two-branch decoder consistently recovers more of the scene geometry, particularly thin structures and distant structures. In cases where a thin structure (e.g. pole) lies close to a larger structure (e.g. walls of building) or a structure is located far away, the single branch decoder often fail to recover their geometry. The two-branch decoder, however, is able to recover distant structures and distinguish thin structures from larger nearby structures. 

\section{Network Architectures}
We trained our model using two architectures: 1) a generic encoder (\tabref{architecture-single-branch}) based on the VGGnet \cite{simonyan2014very} architecture with a single branch decoder 2) a generic encoder (same as the aforementioned) with our proposed two-branch decoder (\tabref{architecture-two-branch}, Sec 4. from main paper).

\begin{table*}[h]
\centering
\scriptsize
\setlength\tabcolsep{15pt}
\begin{tabular}{lccccccc@{}}
                     & \multicolumn{2}{c}{kernel}                  & \multicolumn{2}{c}{channels}                & \multicolumn{2}{c}{downscale}               &                      \\ 
\cmidrule(lr){2-3} 
\cmidrule(lr){4-5}
\cmidrule(lr){6-7}
layer                & size                 & stride               & in                   & out                  & in                   & out                  & input                \\ 
\midrule
Encoder & \multicolumn{1}{l}{} & \multicolumn{1}{l}{} & \multicolumn{1}{l}{} & \multicolumn{1}{l}{} & \multicolumn{1}{l}{} & \multicolumn{1}{l}{} & \multicolumn{1}{l}{} \\
\midrule
conv1                & 7                    & 2                    & 3                    & 32                   & 1                    & 2                    & left                 \\ \midrule
conv1b               & 7                    & 1                    & 32                   & 32                   & 2                    & 2                    & conv1                \\ \midrule
conv2                & 5                    & 2                    & 32                   & 64                   & 2                    & 4                    & conv1b               \\ \midrule
conv2b               & 5                    & 1                    & 64                   & 64                   & 4                    & 4                    & conv2                \\ \midrule
conv3                & 3                    & 2                    & 64                   & 128                  & 4                    & 8                    & conv2b               \\ \midrule
conv3b               & 3                    & 1                    & 128                  & 128                  & 8                    & 8                    & conv3                \\ \midrule
conv4                & 3                    & 2                    & 128                  & 256                  & 8                    & 16                   & conv3b               \\ \midrule
conv4b               & 3                    & 1                    & 256                  & 256                  & 16                   & 16                   & conv4                \\ \midrule
conv5                & 3                    & 2                    & 256                  & 512                  & 16                   & 32                   & conv4b               \\ \midrule
conv5b               & 3                    & 1                    & 512                  & 512                  & 32                   & 32                   & conv5                \\ \midrule
conv6                & 3                    & 2                    & 512                  & 512                  & 32                   & 64                   & conv5b               \\ \midrule
conv6b               & 3                    & 1                    & 512                  & 512                  & 64                   & 64                   & conv6                \\ \midrule
conv7                & 3                    & 2                    & 512                  & 512                  & 64                   & 128                  & conv6b               \\ \midrule
conv7b               & 3                    & 1                    & 512                  & 512                  & 128                  & 128                  & conv7                \\ 
\toprule
Decoder & \multicolumn{1}{l}{} & \multicolumn{1}{l}{} & \multicolumn{1}{l}{} & \multicolumn{1}{l}{} & \multicolumn{1}{l}{} & \multicolumn{1}{l}{} & \multicolumn{1}{l}{} \\ \midrule
upconv7                                  & 3             & 2               & 512          & 512           & 128           & 64            & conv7b                             \\ \midrule
iconv7                                   & 3             & 1               & 1024         & 512           & 64            & 64            & upconv7$\Vert$conv6b                     \\  \midrule
upconv6                                  & 3             & 2               & 512          & 512           & 64            & 32            & iconv7                             \\  \midrule
iconv6                                   & 3             & 1               & 1024         & 512           & 32            & 32            & upconv6$\Vert$conv5b                     \\  \midrule
upconv5                                  & 3             & 2               & 512          & 256           & 32            & 16            & iconv6                             \\  \midrule
iconv5                                   & 3             & 1               & 512          & 256           & 16            & 16            & upconv5$\Vert$conv4b                     \\  \midrule
upconv4                                  & 3             & 2               & 256          & 128           & 16            & 8             & iconv5                             \\  \midrule
iconv4                                   & 3             & 1               & 128          & 128           & 8             & 8             & upconv4$\Vert$conv3b                     \\ \midrule  
\textbf{disp4}                           & 3             & 1               & 128          & 2             & 8             & 8             & iconv4                             \\ \midrule
upconv3                                  & 3             & 2               & 128          & 64            & 8             & 4             & iconv4                             \\ \midrule
iconv3                                   & 3             & 1               & 130          & 64            & 4             & 4             & upconv3$\Vert$conv2b$\Vert$disp4*              \\ \midrule  
\textbf{disp3}                           & 3             & 1               & 64           & 2             & 4             & 4             & iconv3                             \\ \midrule
upconv2                                  & 3             & 2               & 64           & 32            & 4             & 2             & iconv3                             \\ \midrule
iconv2                                   & 3             & 1               & 66           & 32            & 2             & 2             & upconv2$\Vert$conv1b$\Vert$disp3*              \\ \midrule  
\textbf{disp2}                           & 3             & 1               & 32           & 2             & 2             & 2             & iconv2                             \\ \midrule
upconv1                                  & 3             & 2               & 32           & 16            & 2             & 1             & iconv2                             \\ \midrule
iconv1                                   & 3             & 1               & 18           & 16            & 1             & 1             & upconv1$\Vert$disp2*                     \\ \midrule
\textbf{disp1}                           & 3             & 1               & 16           & 2             & 1             & 1             & iconv1                             \\ \midrule

\end{tabular}
\vspace{1em}
\caption{Our network architecture follows that of \cite{godard2017unsupervised} and \cite{mayer2016large} and we are able to outperform the baseline \cite{godard2017unsupervised}. ``in'' and ``out'' refers to the input and output channels and downscale factor due to striding for each layer. $\Vert$ refers to the concatenation of multiple layers. $*$ refers to up-sampling disparity predictions at a given resolution. Batch normalization was not used.}
\label{architecture-single-branch}
\end{table*}

\twocolumn

\begin{table}[!ht]
\begin{adjustwidth}{-.2in}{0.2in} 
\centering
\scriptsize
\begin{tabularx}{.47\textwidth}{p{0.8cm}ccccccc} 
                     & \multicolumn{2}{c}{kernel}                  & \multicolumn{2}{c}{channels}                & \multicolumn{2}{c}{downscale}               &                      \\ 
\cmidrule(lr){2-3} 
\cmidrule(lr){4-5}
\cmidrule(lr){6-7}
layer                & size                 & stride               & in                   & out                  & in                   & out                  & input                \\ 
\midrule
Encoder & \multicolumn{1}{l}{} & \multicolumn{1}{l}{} & \multicolumn{1}{l}{} & \multicolumn{1}{l}{} & \multicolumn{1}{l}{} & \multicolumn{1}{l}{} & \multicolumn{1}{l}{} \\
\midrule
conv0                & 7                    & 1                    & 3                    & 32                   & 1                    & 1                    & left                 \\ \midrule
conv1                & 7                    & 2                    & 32                   & 32                   & 1                    & 2                    & conv0                \\ \midrule
conv1b               & 7                    & 1                    & 32                   & 32                   & 2                    & 2                    & conv1                \\ \midrule
conv2                & 5                    & 2                    & 32                   & 64                   & 2                    & 4                    & conv1b               \\ \midrule
conv2b               & 5                    & 1                    & 64                   & 64                   & 4                    & 4                    & conv2                \\ \midrule
conv3                & 3                    & 2                    & 64                   & 128                  & 4                    & 8                    & conv2b               \\ \midrule
conv3b               & 3                    & 1                    & 128                  & 128                  & 8                    & 8                    & conv3                \\ \midrule
conv4                & 3                    & 2                    & 128                  & 256                  & 8                    & 16                   & conv3b               \\ \midrule
conv4b               & 3                    & 1                    & 256                  & 256                  & 16                   & 16                   & conv4                \\ \midrule
conv5                & 3                    & 2                    & 256                  & 512                  & 16                   & 32                   & conv4b               \\ \midrule
conv5b               & 3                    & 1                    & 512                  & 512                  & 32                   & 32                   & conv5                \\ \midrule
conv6                & 3                    & 2                    & 512                  & 512                  & 32                   & 64                   & conv5b               \\ \midrule
conv6b               & 3                    & 1                    & 512                  & 512                  & 64                   & 64                   & conv6                \\ 
\bottomrule

\end{tabularx}
\vspace{1em}
\caption{Our proposed network architecture, which achieves state-of-the-art. ``in'' and ``out'' refers to the input and output channels and downscale factor due to striding for each layer. $\Vert$ refers to the concatenation of multiple layers. $*$ refers to up-sampling disparity predictions at a given resolution. The branch prefixed with `\texttt{i}' makes the initial prediction and the branch prefixed with `\texttt{r}' makes the final prediction.}
\label{architecture-two-branch}
\end{adjustwidth} 
\end{table}
\newpage
\begin{table}[!h]
\begin{adjustwidth}{-.45in}{-0.1in} 
\centering
\scriptsize
\vspace{-0.2em}
\begin{tabularx}{.59\textwidth}{p{0.6cm}ccccccc}
                     & \multicolumn{2}{c}{kernel}                  & \multicolumn{2}{c}{channels}                & \multicolumn{2}{c}{downscale}               &                      \\ 
\cmidrule(lr){2-3} 
\cmidrule(lr){4-5}
\cmidrule(lr){6-7}
layer                & size                 & stride               & in                   & out                  & in                   & out                  & input                \\ 
\midrule
Decoder & \multicolumn{1}{l}{} & \multicolumn{1}{l}{} & \multicolumn{1}{l}{} & \multicolumn{1}{l}{} & \multicolumn{1}{l}{} & \multicolumn{1}{l}{} & \multicolumn{1}{l}{} \\ \midrule
iupconv6                                  & 3             & 2               & 512          & 512          & 64            & 32            & conv6b                             \\  \midrule
iconv6                                    & 3             & 1               & 1024         & 512          & 32            & 32            & iupconv6$\Vert$conv5b              \\  \midrule
iupconv5                                  & 3             & 2               & 512          & 256          & 32            & 16            & iconv6                             \\  \midrule
iconv5                                    & 3             & 1               & 512          & 256          & 16            & 16            & iupconv5$\Vert$conv4b              \\  \midrule
iupconv4                                  & 3             & 2               & 256          & 128          & 16            & 8             & iconv5                             \\  \midrule
iconv4                                    & 3             & 1               & 256          & 128          & 8             & 8             & iupconv4$\Vert$conv3b              \\ \midrule  
idisp4                                    & 3             & 1               & 128          & 2            & 8             & 8             & iconv4                             \\ \midrule
sconv4                                    & 3             & 1               & 128          & 128          & 8             & 8             & conv3b                             \\ \midrule
sconv4b                                   & 3             & 1               & 128          & 128          & 8             & 8             & sconv4                             \\ \midrule
rskip4                                    & 3             & 1               & 128          & 128          & 8             & 8             & conv3b+sconv4b                     \\ \midrule
rconv4                                    & 3             & 1               & 258          & 128          & 8             & 8             & iconv4$\Vert$idisp4$\Vert$rskip4   \\ \midrule
\textbf{rdisp4}                           & 3             & 1               & 128          & 2            & 8             & 8             & rconv4                             \\ \midrule
iupconv3                                  & 3             & 2               & 128          & 64           & 8             & 4             & iconv4                             \\ \midrule
iconv3                                    & 3             & 1               & 130          & 64           & 4             & 4             & iupconv3$\Vert$conv2b$\Vert$idisp4*  \\ \midrule 
idisp3                                    & 3             & 1               & 64           & 2            & 4             & 4             & iconv3                             \\ \midrule
sconv3                                    & 3             & 1               & 64           & 64           & 4             & 4             & conv2b                             \\ \midrule
sconv3b                                   & 3             & 1               & 64           & 64           & 4             & 4             & sconv3                             \\ \midrule
rskip3                                    & 3             & 1               & 64           & 64           & 4             & 4             & conv2b+sconv3b                     \\ \midrule
rupconv3                                  & 3             & 2               & 128          & 64           & 8             & 4            & rconv4                             \\ \midrule
rconv3                                    & 3             & 1               & 196          & 64           & 4             & 4             & iconv3$\Vert$idisp3$\Vert$rupconv3$\Vert$rskip3$\Vert$rdisp4*   \\ \midrule
\textbf{rdisp3}                           & 3             & 1               & 64           & 2            & 4             & 4             & rconv3                             \\ \midrule
iupconv2                                  & 3             & 2               & 64           & 32           & 4             & 2             & iconv3                             \\ \midrule
iconv2                                    & 3             & 1               & 66           & 32           & 2             & 2             & iupconv2$\Vert$conv1b$\Vert$idisp3*              \\ \midrule  
idisp2                                    & 3             & 1               & 32           & 2            & 2             & 2             & iconv2                             \\ \midrule
sconv2                                    & 3             & 1               & 32           & 32           & 2             & 2             & conv1b                             \\ \midrule
sconv2b                                   & 3             & 1               & 32           & 32           & 2             & 2             & sconv2                             \\ \midrule
rskip2                                    & 3             & 1               & 32           & 32           & 2             & 2             & conv1b+sconv2b                     \\ \midrule
rupconv2                                  & 3             & 2               & 64           & 32           & 4             & 2             & rconv3                             \\ \midrule
rconv2                                    & 3             & 1               & 100          & 32           & 2             & 2             & iconv2$\Vert$idisp2$\Vert$rupconv2$\Vert$rskip2$\Vert$rdisp3*   \\ \midrule
\textbf{rdisp2}                           & 3             & 1               & 32           & 2            & 2             & 2             & rconv2                             \\ \midrule
iupconv1                                  & 3             & 2               & 32           & 16           & 2             & 1             & iconv2                             \\ \midrule
iconv1                                    & 3             & 1               & 18           & 16           & 1             & 1             & iupconv1$\Vert$idisp2*             \\ \midrule
idisp1                                    & 3             & 1               & 16           & 2            & 1             & 1             & iconv1                             \\ \midrule
sconv1                                    & 3             & 1               & 32           & 32           & 1             & 1             & conv0                              \\ \midrule
sconv1b                                  & 3              & 1               & 32           & 32           & 1             & 1             & sconv1                             \\ \midrule
rskip1                                   & 3             & 1                & 32           & 32           & 1             & 1             & conv0+sconv1b                      \\ \midrule
rupconv1                                 & 3             & 2                & 64           & 32           & 2             & 1             & rconv2                             \\ \midrule
rconv1                                   & 3             & 1                & 68           & 16           & 1             & 1             & iconv1$\Vert$idisp1$\Vert$rupconv1$\Vert$rskip1$\Vert$rdisp2*   \\ \midrule
\textbf{rdisp1}                          & 5             & 1                & 16           & 2            & 1             & 1             & rconv1                             \\ \midrule
\end{tabularx}
\end{adjustwidth}
\end{table}

\end{document}